\documentclass{article}

\usepackage{PRIMEarxiv}

\usepackage[utf8]{inputenc} 
\usepackage[T1]{fontenc}    
\usepackage{hyperref}       
\usepackage{url}            
\usepackage{booktabs}       
\usepackage{amsfonts}       
\usepackage{nicefrac}       
\usepackage{microtype}      
\usepackage{lipsum}
\usepackage{fancyhdr}       
\usepackage{graphicx}       
\graphicspath{{media/}}     

\title{A Critical Analysis on Machine Learning Techniques for Video-based Human Activity Recognition of Surveillance Systems: A Review
}

\author{
  Shahriar Jahan, Roknuzzaman, Md Robiul Islam\\
  Department of Mechatronics Engineering \\
  Rajshahi University of Engineering \& Technology \\
  Rajshahi-6204, Bangladesh\\
  \texttt{shahriar.sakib.glhs@gmail.com, lara05ll95@gmail.com, robiulislamme07@gmail.com} 
}

\begin{document}
\maketitle

\begin{abstract}
Upsurging abnormal activities in crowded locations such as airports, train stations, bus stops, shopping malls, etc., urges the necessity for an intelligent surveillance system. An intelligent surveillance system can differentiate between normal and suspicious activities from real-time video analysis that will enable to take appropriate measures regarding the level of an anomaly instantaneously and efficiently. Video-based human activity recognition has intrigued many researchers with its pressing issues and a variety of applications ranging from simple hand gesture recognition to crucial behavior recognition in a surveillance system. This paper provides a critical survey of video-based Human Activity Recognition (HAR) techniques beginning with an examination of basic approaches for detecting and recognizing suspicious behavior followed by a critical analysis of machine learning and deep learning techniques such as Convolutional Neural Network (CNN), Recurrent Neural Network (RNN), Hidden Markov Model (HMM), K-means Clustering etc. A detailed investigation and comparison are done on these learning techniques on the basis of feature extraction techniques, parameter initialization, and optimization algorithms, accuracy, etc. The purpose of this review is to prioritize positive schemes and to assist researchers with emerging advancements in this field's future endeavors. This paper also pragmatically discusses existing challenges in the field of HAR and examines the prospects in the field. 
\end{abstract}


\section{Introduction}
There have been alarming rates of crime of different kinds around the world. The manual surveillance system which depends on the human security guards demands more effort and financial expenses. Thus the existing surveillance systems have lost their credibility and therefore, arise the need for an artificial intelligence-based video surveillance system \cite{gowsikhaa2014automated} and that requires Human Activity Recognition (HAR). Video-based HAR can generally be classified into two groups depending on motion feature, i.e. marker-based and vision-based where the first-mentioned refers to as optical marker-based motion capture (MoCap) system, and the second-mentioned is formed on depth video cameras which are marker-free. Because of that particular reason, the latter can be used for HAR in the surveillance system. In recent years, recognition using Deep Learning (DL) has gained tremendous prominence because of its ability to learn deep structures of patterns. DL methods are becoming more and more highly regarded each day and gaining significant attention from pattern recognition and application researchers. As a result in the recent past, more research funding has been assigned to the particular sector. Conceptually, deep learning alludes to neural networks that utilize layers of non-linear data processing for feature classification \cite{park2016depth}. Layers are structured hierarchically together with outcomes for preceding layers are processed by current layers. Within the realm of computer vision, DL approaches have surpassed many conventional methods \cite{krizhevsky2017imagenet,simonyan2014very,szegedy2015going,duffner20143d}.
\par
\par

 An intelligent surveillance system monitors vulnerable areas such as ATMs and banks for detecting violence, burglary, fire, and fighting-like incidents \cite{gowsikhaa2014automated}. The main objective of a deep learning-based video surveillance system is to specify suspicious behavior from the captured video from certain areas where the abnormality of behavior can be categorized based on activity detection, feature extraction, classification, and recognition of the interesting patterns \cite{park2016depth,lavee2007framework}. This process has been an arising research issue as to perpetuate surveillance in an encapsulated video that contains detecting, classifying, tracking, and recognizing the doings of a human.
 Not having HAR has led to too many incidents turn into accidents. A few of the accidents that frequently happen which can be prevented by having a HAR system are listed below.
 \begin{enumerate}
 
     \item Bank robbery has been more frequent these days than in the last few decades. Lack of smart surveillance \cite{kakadiya2019ai} the robbers often get away before the police arrive or get information about the bank getting robbed. 
     \item Sudden Cardiac Death (SCD) \cite{markwerth2021sudden} is a common phenomena happening all around the world. Most cases of death have happened not getting the treatment timely or not reacting for the proper step which is to be taken.
     \item ATM robbery \cite{viji2021intelligent} is a common scene nowadays. ATM booths all over the world are prone to severe criminal activities. In any case of criminal activity, the victims become helpless when the booths are located in remote areas. The investigation officials are aware of the crime activity only after the crime is over after the robbers escape.
     
 \end{enumerate}

 \par
 Therefore, several processes have been outlined and implemented by many researchers but flawless intelligent surveillance has not been developed yet \cite{verma2019review}.
 \par
 \par

\begin{figure}[hbt!]
  \centering
\includegraphics[scale=.55]{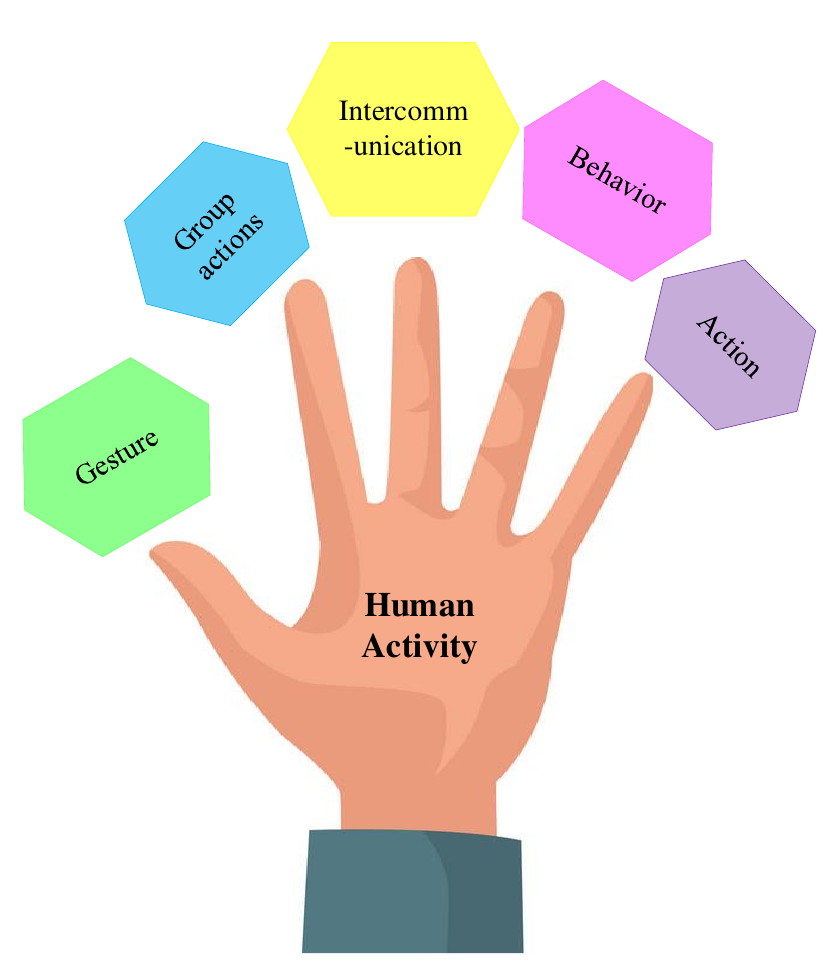}
\caption{Partition of Human Action \cite{vrigkas2015review}}
\label{fig:2}
\end{figure}

Human action recognition expects to affirm human behaviors from a video. To precisely group, input information into its covered-up category, and to successfully characterize input data into its hidden movement category is the principal objective of the human action recognition system. Contingent upon their complexity, human exercises are classified into:  
atomic actions, gestures, intercommunication, group actions, and behaviors. Fig. \ref{fig:2} explains the partition of human action as maintained by the complexities. Gestures are considered as crude developments of the body portions of an individual that may compare to a specific activity of an individual \cite{yang2012discovering}. Atomic activities are developments of an individual portraying a specific movement that might be a piece of progressively complex exercises \cite{ni2015motion}. Intercommunication between people is a human exercise that includes at least two people \cite{patron2012structured}. Group activities are exercises executed by a gathering of individuals \cite{tran2014activity}. Human behavior alludes to physical activities that are related to the feelings, character, and mental conditions of a person.

\begin{figure}[hbt!]
  \centering
  
\includegraphics[scale=.30]{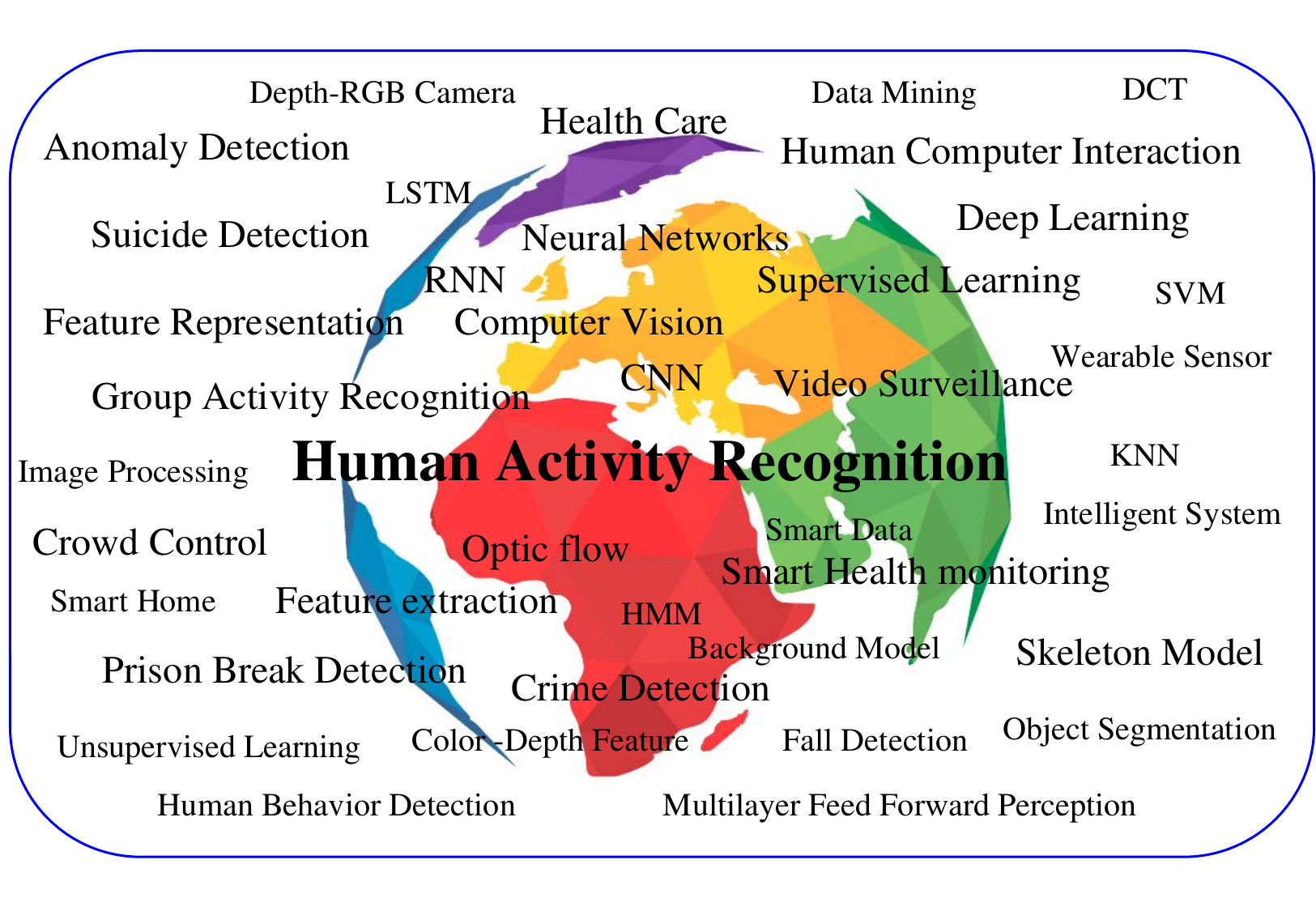}
\caption{Keywords used in many research articles related to HAR}
\label{keywords}
\end{figure}

The eyes of the video monitoring system are cameras. Cameras may be mounted in ideal positions and be dynamic in motion to help display items without obstruction. Internet protocol (IP) cameras are used for high-quality video applications. For human behavior analysis, video analysis is used. The absence of static images will maximize storage as saving all videos requires a lot of memories. This is achieved if a video recording sequence is activated when movement happens in a scene hence storage costs are reduced \cite{gowsikhaa2014automated}. A human being can identify things in motion intuitively only by watching a video. But the video must be separated into a series of images first to make a computer do so. A task consisting of three elements required to overcome the issue are background subtraction \cite{elgammal2002background}, human tracking \cite{wang2013dense} and human action recognition \cite{gan2015devnet}. Fig. \ref{keywords} focuses most of the used keywords in the research field of HAR. In background separation, the system attempts to differentiate among different pieces of an image that are time-invariant. In human tracking, the system distinguishes human movement over time and motion detection helps a system to detect the video part. In human action recognition, the system confines a human movement in a motionless picture. The initial steps involved in processing a video in the HAR system consist of preprocessing, motion detection, motion tracking, feature extraction, object classification, and finally, activity recognition as shown in Fig. \ref{fig:1}. The motion detection module consists of background modeling, foreground modeling, and foreground segmentation. Background modeling derives a background image from a frame-set. Modeling parts of the foreground puts out a vivid picture of moving objects. The segmentation of the foreground removes noise and leads to a blob representing the silhouette of the human.
\par

Two of the common problems in the video remain among different methods of classification, the recognition problem and the localization problem \cite{vrigkas2015review}. One must decide the active conditions of an individual to let the system run efficiently When endeavoring to perceive human activities. Activities, for example, "handshaking" and "walking," emerge very normally in everyday life and are moderately simple to perceive. Again, increasingly complex exercises, for example, "opening a bottle", "wearing a wristwatch" are increasingly hard to identify. The improvement of a completely automated human movement detection framework, fit for ordering an individual's movements with a low mistake, is a difficult assignment because of changes in scale, background distortion, partial occlusion, frame resolution, viewpoint, and lighting.

In the last decade, a large amount of research has been conducted in the field of HAR. Table 1 represents that this paper highlights various phases of video processing techniques and provides a detailed overview and comparison of several of the most used supervised and unsupervised learning techniques and identified their limitations, result parameters, and proposed some future trends to work on, in the field of human activity recognition. The research also differentiates based on the adopted techniques of each algorithm. In this research different stages of a complete activity recognition system are separately discussed. Data preprocessing and feature extraction methods of different machine learning techniques applied in numerous previous works are briefly reviewed. Different video-based adopted features are used in different supervised and unsupervised learning techniques and the most used human action datasets are also been explored in this research. On the other hand, various similar review papers failed to cover most of these aspects and lack depth of discussion and comparison. In \cite{ann2014human}, the authors provided a review on HAR sensing technologies where lacking a thorough discussion on the feature extraction or the learning models have been noticed. Also, the research does not contain any comparison of the result parameters of the works they reviewed. \cite{ke2013review} reviewed on video-based HAR but it also lacks comparison regarding the result parameters or any clear distinction between various approaches of the reviewed papers. Moreover, they only discussed stochastic and segmentation-based models ignoring the state of art deep learning approaches. \cite{verma2019review} reviewed a few supervised learning and unsupervised machine learning models but did not go through full depths and also did not go through the state-of-the-art deep learning models. \cite{dhiman2019review} surveyed the state of the art techniques of HAR but the emphasis was put on feature representations of actions rather than the learning methods. The discussion on datasets is also poorly represented.
\par

\begin{table*}[h]
\centering
	\caption{Keywords used in many research articles related to HAR.}
	\includegraphics[width=0.98\textwidth]{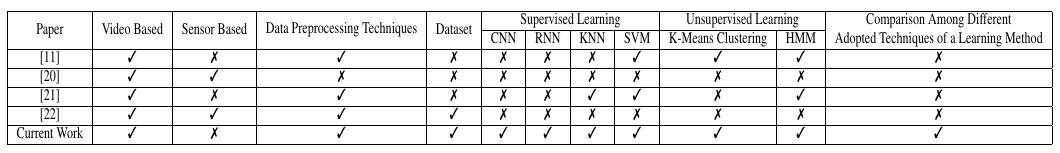}
		\label{fig:data}
\end{table*}

\subsection{Paper Organization}
The paper begins by explaining the need for HAR based surveillance system with a glimpse of how the system works. The outline of the paper is arranged as follows: A basic approach towards HAR is discussed in section II. A brief overview of supervised and supervised learning methods used in HAR is in sections III and IV consecutively. After that a detail of datasets used in HAR is discussed in section V . The challenges and future trends of the reviewed learning methods for surveillance systems are discussed in section VI with the conclusion of the work in section VII.

\section{Basic approach towards HAR}
\begin{figure}[hbt!]
  \centering
\includegraphics[scale=.45]{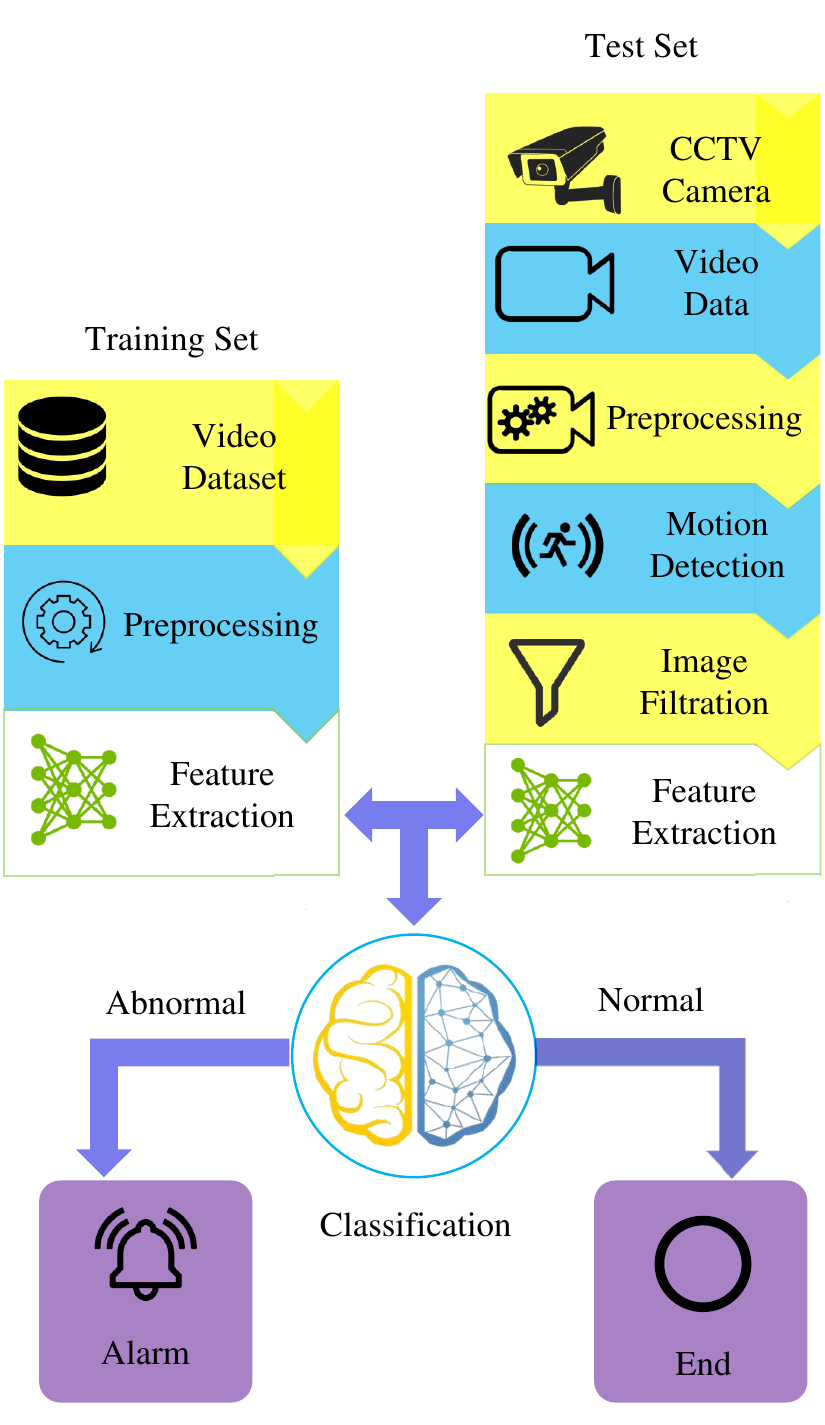}
 \caption{Basic model of activity detection and classification in intelligent surveillance system \cite{ann2014human}\cite{alruwaili2019human}}
  \label{fig:1}
\end{figure}

In Fig. \ref{fig:1} the basic model towards HAR is shown. This model has already been used in past works of literature. This model contains important steps like preprocessing, motion detection, data filtration, feature extraction. This methodology has an advantage that is all the supervised machine learning algorithms apply it.

\subsection{Preprocessing}
Data preprocessing is an indispensable stage in Artificial Intelligence \cite{yang2018data}. The main intention is to get the final dataset which is acknowledged as suitable and helpful data for algorithms \cite{garcia2016tutorial}. It was suggested by Pyle \cite{pyle1999data} that more than $50\%$ of executed time should be spent on preprocessing. It also has a significant impression on the quality and quantity of the identified knowledge during the process \cite{acuna2011preprocessing}.
The real-world dataset, associated with human activity, is extremely sensitive to noise, mechanical failures, missing including irregular data because of human faults, and to their typically large quantity which is known as dirty data. Data quality elements such as accuracy, completeness, consistency, believability, and interpretability get hampered without preprocessing. Believability resembles how much the user trusts the data and interpretability means how much the data is understandable \cite{han2011data}. Improvement of the data quality depends on applying suitable preprocessing techniques. For that purpose of helping to enhance the performance of the following process, \cite{lu1996preprocessing},\cite{pyle1999data} and \cite{azzopardi2002right} suggested representations of established methods for outfitting data for analysis. The methods delineated by the authors are summed as follows:

\subsubsection{Data Cleaning}
The data cleaning techniques have been formed and executed to refine data for data exploration and analysis. It is done by filling in lost values or value ascription, grading data containing noise, classifying and/or excreting outliers, and correcting inequalities \cite{garcia2016tutorial}. Smoothing methods that are associated with discretization are likewise known as methods of data reduction. The reason behind this is it lessens the number of separated values per attribute. For Detecting outliers to remove noise, Clustering methods can also be used \cite{garcia2016tutorial}. Overall the main task of cleaning is filling in absent data, smoothing data consisting of noise, passing anomalies, and fixing inequalities.

\subsubsection{Data Integration}
Many systems demand the integration of collective database or multiple records \cite{acuna2011preprocessing}. A system similar to this phenomenon requires data integration as preprocessing task. Data integration merges data that are associated with the system to produce more consistent, accurate, and more useful information. From multiple real-world data sets, it takes system-oriented data and merges them into one without harming or deflecting the system result. Conventional processes achieved within the data integration are the classification and union of variables as well as domains, the study of characteristics, the duplication of tuples, and the exposure of conflicts in data values of separate sources \cite{garcia2016tutorial}.

\subsubsection{Data Transformation}
Data transformation is a preprocessing step that transforms and/or consolidates data from real-world data. It generally makes the process result suitable for the system and makes it more efficient. The task is divided into several sub-tasks that are feature construction, smoothing, normalization, aggregation, and generalization of data \cite{garcia2015data}.
\par

\subsubsection{Data Characterization}
Data characterization details data on how a system or a user demands. \cite{engels1998using} outlined the following qualities that are considered as ideal for a provided dataset and they are the number of classes, the number of checks, the percentage of missing contents in each attribute, the number of qualities, the number of features with an integer data type as well as the number of features with a symbolic type of data \cite{acuna2011preprocessing}. Location parameters are quantification for example maximum, minimum, average, median, and practical quartiles although dispersion criteria such as range, standard departure, and quartile deviation render quantification that specifies the scattering contents of the class \cite{garcia2015data}. Location plus dispersion specifications are classified into two categories. One is the kind that can work with extreme values and the other is the kind that is sensitive to extreme values \cite{garcia2016tutorial}.

\subsubsection{Data Visualization}
For the data exploration and characterization phase, having the assistance of data Visualization methods can similarly be very helpful. Visualizing data before preprocessing helps to understand the data \cite{garcia2016tutorial}. Because of that novel and valuable knowledge can be deduced from the data. To distinguish the presence of absent contents, and anomalies, as well as to recognize correlations among characteristics, visualization methods conceivably be used which helps ranking data according to contaminants and to select the most appropriate data for the system \cite{acuna2011preprocessing}.

\subsection{Feature extraction}
In the identification of patterns and pictures methods of finding a few significant points or identification points of interest that can describe the quality of the images as boundaries, 
Corners, cloth and so on. Feature extraction is a technique for dimension reduction where massive data are modified to features, which are designated as vectors since the larger length of the input has more data size but fewer facts.
\par
\textit{Data reduction} produces data that represents or replicates the raw data or real-world data but in a much smaller volume which produces the same analytical result. The reduced data provides more efficiency yet delivering the same result.
\par
Data reduction is performed utilizing feature selection and/or instance selection. Feature selection seeks to classify the most relevant, explanatory input variables within a data set.
\cite{han2011data} suggested the succeeding approaches for data reduction:
\par
\textit{Dimension reduction} is known as the process of lessening the number of arbitrary variables or traits or features under consideration. It is a data reduction task where algorithms are employed to erase unrelated and unneeded attributes \cite{han2011data}.
\par
\textit{Data compression} is the data reduction process where transformations are employed to achieve a decreased or compressed rendering or replication of the real data. Two common categories are wavelet transforms and Principal Component Analysis (PCA) \cite{basly2021dtr}. If the original data can be regenerated without any data loss from the compressed data then it is called lossless. If not or if there can be generated a quite approximate data that has a number of data losses then it is called lossy.
\par
\textit{Numerosity reduction} is the reduction process in which the data are restored or evaluated by stand-in compact data characterization acting as parametric models that just keep the paradigm specifications or non-parametric techniques for example clustering, sampling, data cube aggregation, and the application of histograms are mostly used.
\par
There exist many other methods to organize the data reduction process. The computational time estimated to spend should not be outweighed. 
\subsection{Classification}
Classification is the method of classifying the data sequences. The input data is classified in class numbers. This is normally applied to allocate freshly tested data according to the learned model into class tags. Normal and abnormal incidents are typically a question with two classes. Going a bit deeper, abnormal cases can also be classified into different levels depending on the actions of a human in a time sequence. As more complex cases arise in order to achieve human-level performance, the algorithms had to be involved. 

While HAR has been an active research topic for quite a long time, traditional HAR algorithms have struggled in the past with some critical problems. Some of the major problems were:

i. Previous HAR-based methods depended primarily on time series data where most of the features had to be hand-crafted for the AI algorithms to get a good enough accuracy. The hand-crafting feature was often a lengthy and time-consuming process.

ii. Another problem of hand-crafted feature-based models is that these relied on a lot of factors and were not a good fit whenever subtle things such as camera movement, harsh environments such as rainy or smokey situations aroused. Because of this, these models often only worked on the datasets they were trained for, which was quite impractical.

iii. Traditional approaches had to go through a lot of steps while detecting abnormal behavior, where the recent approaches use end-to-end models.

iv. Traditional approaches were not accurate enough with image data, hence vision-based HAR approaches were not a viable solution until state-of-the-art deep learning solution came into the scenario which seemed to have remarkable performance in working with images.

\begin{figure}[hbt!]
  \centering
\includegraphics[scale=.345]{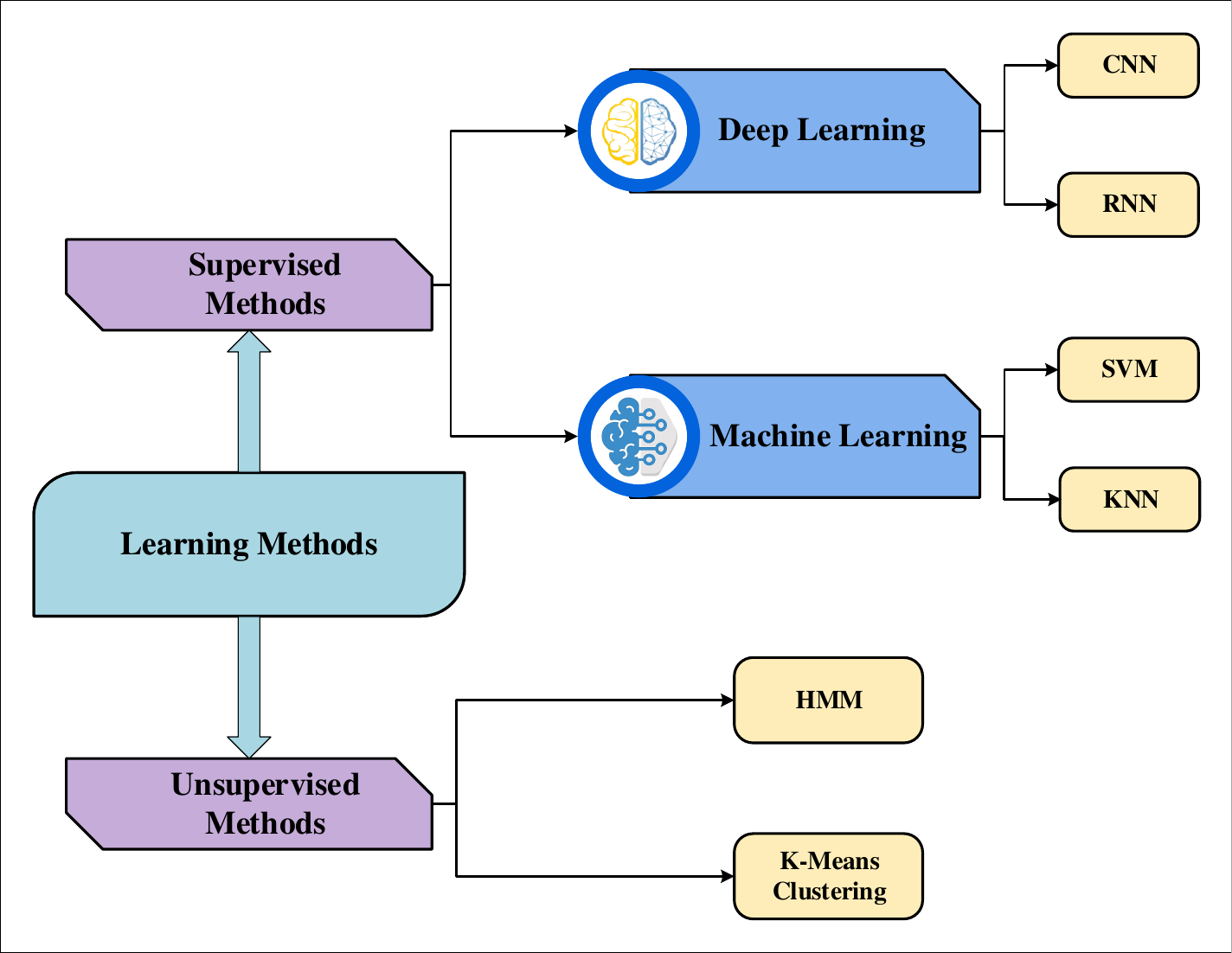}
\caption{Tree diagram of the learning methods discussed in this paper }
\label{ML}
\end{figure}

Many researchers used different types of classifiers to identify usual and anomalous events. K-Nearest Neighbour (KNN), Support Vector Machine (SVM), Hidden Markov model (HMM) are some of the most used machine learning algorithms used in this research sector. Also, some of the state-pf-the-art supervised deep learning algorithms such as  Convolutional Neural Network (CNN) \cite{fahim2020visual}, Recurrent Neural Network (RNN), etc. have also been used. All of them are discussed in the following section. Fig. \ref{ML} shows a tree diagram of different learning methods discussed in this paper. 

\section{Anomaly Detection using Supervised Learning}

An anomaly can be defined as a sequence of actions that can be used to determine whether the activity is abnormal in the context of the environment, place, or social standard. For example, people running in a marathon may be determined as a normal activity but people running in the middle of the street at an odd time may be considered an abnormal activity. Anomaly detection can be crucial in several cases such as:

i. Sudden crowd gatherings for doing any mischievous activities.

ii. Sudden criminal activities like people fighting, or a person getting mugged/ kidnapped/ attacked by another person. 

iii. People suddenly falling off in the street which may be a case of a heart attack or other disease.

iv. Sudden road accidents. 
\begin{figure*}[hbt!]
  \centering
\includegraphics[scale=.47]{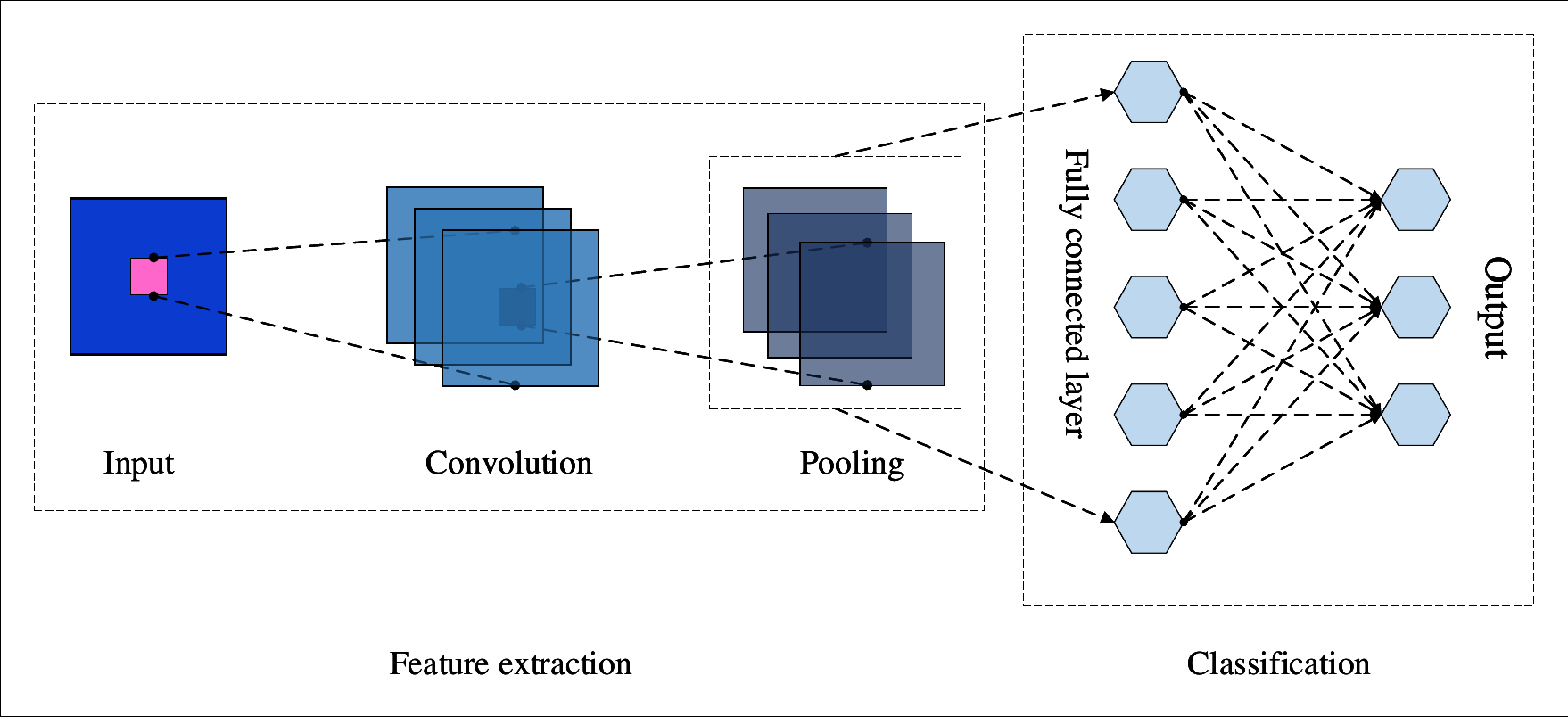}
  \caption{Basic Architecture of CNN}
\label{CNN1}
\end{figure*}
Supervised learning is the method of learning in which a model is trained initially with the required dataset and after that classifies the new data given to it as an input into normal and anomaly \cite{verma2019review}. Development in the field of HAR and anomaly detection for surveillance using supervised learning, especially deep learning methods has proven to be ground-breaking \cite{mohan2019anomaly}. Supervised learning conceivably is severed into an individual class or multi-class problems. Thus some important classifiers for anomaly detection in HAR are discussed in this section.

\subsection{Convolutional Neural Network} 

Convolutional Neural Networks (CNNs) have demonstrated their superiority in many applications \cite{sarker2019multidimensional,fahim2020self,basly2021dtr}. Among them, the development of HAR systems has been greatly benefited by CNN variants. Being in the class of deep neural networks, CNN is used to peruse visual entities \cite{agarwal2012simple}. It employs a mathematical method named convolution operation thus named Convolutional neural network. CNN resembles more like a neural network possessing a grid-like topology in the process \cite{lecun1989generalization,fahim2020self}. For a neuro-scientiﬁc principle inﬂuencing deep learning, CNN stands out of all other algorithms. 


CNN comprises two parts, the hierarchical feature extractor that holds the convolutional with the pooling layers, and the fully connected layer which classifies the feature vectors \cite{sun2018sequential}.
In the fully connected layer, hidden layers typically incorporate a sequence of convolutional layers which convolute with multiplication operations. The convolved output is passed consequently to the next layer. Fig. \ref{CNN1} shows a basic layer terminology of CNN. 
The lower layers of CNN  undertake the local salient data and the higher layers undertake salient patterns from the data. However, multiple salient patterns jointly affect and resembles the output as CNN is a multilayered network \cite{yang2015deep,sarker2020regularized}.

Manual feature extraction, as well as filtering, is not essential in CNN-based activity recognition. CNN can easily obtain complex gestures of humans while investigating nonlinear as well as time-domain interconnections \cite{lee2019optimal}. Three important ideas make CNN advantageous sparse interaction, equivariant representation, and parameter sharing \cite{Goodfellow-et-al-2016}. Space interaction is gained by declaring the kernel smaller than the input that helps to save both memory and run time. Equivariance to translation in convolution resembles the specific structure of parameter sharing \cite{kaghyan2012activity}.
 \cite{sheng2016short} employed CNN with two convolutional and pooling layers that led them to result in a recognition rate of $95.9\%$ in human action recognition.

\begin{table*}[h]
\centering
	\caption{Comparison between different features of CNN in HAR.}
	\includegraphics[width=0.98\textwidth]{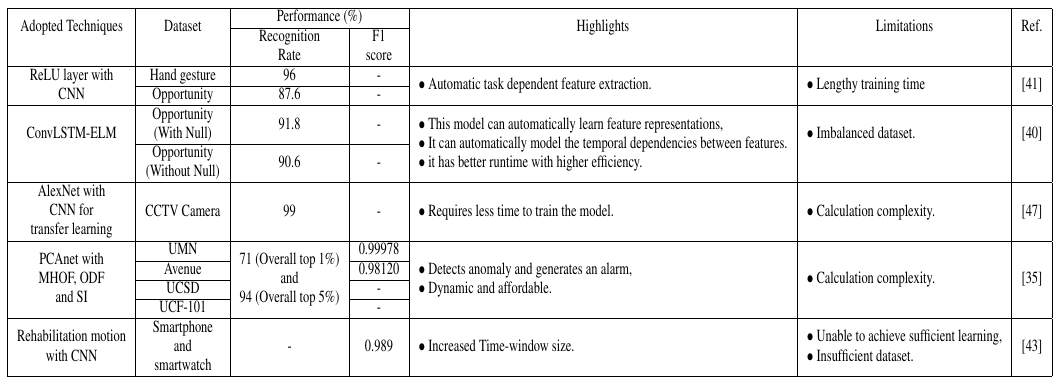}
		\label{fig:data}
\end{table*}

\begin{figure*}[hbt!]
  \centering
\includegraphics[scale=.80]{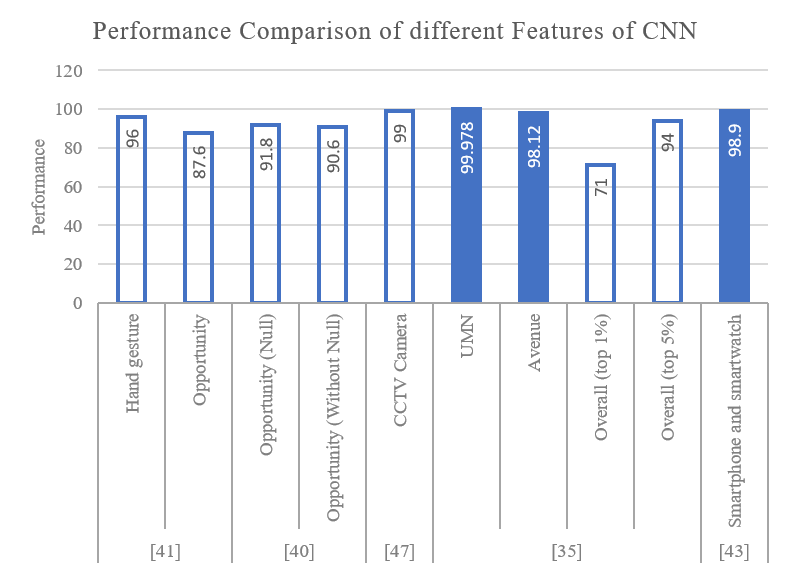}
  \caption{Histogram of the performance comparison of different features of CNN}
\label{CNNchart}
\end{figure*}

Among the three stages in convolution, the second stage linear activation has to pervade in some nonlinear activation function which is the detector stage \cite{Goodfellow-et-al-2016}. In the next stage, pooling is employed for the activation function to be linearized once again. The pooling layer decreases data dimension by mixing outcomes of neuron clusters which are at the current layer into a solitary neuron in the following one. The local pooling layer combines small layers where the global pooling layer works on all the neurons in the convolution \cite{ciresan2011flexible}.
Pooling can calculate the maximum value as well as the average value. Max pooling \cite{sarker2019multidimensional} operation uses maximum values which got from each of the clusters used \cite{ciregan2012multi}, \cite{zhou1988computation}. On the other hand, the average pooling operation uses the mean values got from each of the clusters used \cite{mittal2018survey}. Both maximum pooling and average pooling works with a cluster of neurons at the prior layer.
\par
In the paper \cite{anishchenko2018machine} transfer learning technique is applied by using pre-trained CNN AlexNet \cite{krizhevsky2017imagenet} for abnormal activity recognition. This technique transfers knowledge gained by solving one task to solve another. Therefore, it requires less time training CNN model for new datasets compared to train it from the scratch. The author made changes to layers of the already existing CNN AlexNet architecture to distinguish the activities into two classes which are required for abnormal activity recognition. In \cite{yang2015deep}, the authors constructed the CNN layer architecture in groups of five sections. In each of the first three sections incorporation of normalization layer and Rectified Linear Unit (ReLU) layer had been done to gain better results.
\par 
\cite{sun2018sequential} proposed a hybrid architecture for human activity recognition where feature extraction and classification are not solely dependent on convolutional layers like the traditional algorithms. The hybrid architecture comprises convolutional layers for feature extraction, Long Short-Term Memory (LSTM) layers for producing temporal dependencies of the extracted features by convolutional layers, and an Extreme Learning Machine (ELM) classifier to replace the fully connected layers working on abnormal human activity classification. The main challenges faced by researchers in HAR are large varieties of classes including a massive proportion of Null classes and the similarities between the classes. Analyzing the activities and putting them into proper classes requires an enormous amount of time. 
\par
A model proposed in \cite{ordonez2016deep} exploiting CNN and LSTM solved this problem but real-time performance could not be achieved. Inclusion of ELM as a classifier immensely improved real-time performance that has been shown in \cite{sun2018sequential}.
\par 
In \cite{mohan2019anomaly}, the authors proposed A CNN model in which at first abnormal frames will be detected then the abnormal activity in that frame will be located. Features of Saliency Information (SI) and Multi-scale Histogram of Optical Flow (MHOF) have been applied for recognizing abnormal frames. For reducing computational complexity MHOF divides each frame into many image patches and then processes them. MHOF can detect anomalies in real-time with high precision. The Principal Component Analysis (PCA) is used for higher-order extraction from the previously extracted features and those higher-level features have been trained in SVM. CNN (frame-wise classification) has an accuracy of $69\%$ and $92\%$ respectively in top 1 and top 5 conditions. When CNN to extract features with Recurrent Neural Network(RNN) has an accuracy of $71\%$ and $94\%$ respectively in top 1 and top 5 conditions. A compact representation of all the features discussed in this sub-section is compared in Table 2. Fig. \ref{CNNchart} Shows a histogram of comparison among different features of CNN where the filled pillars represent f1 score $\times 100\%$ and the hollow pillars represent recognition rate.

\subsection{Recurrent Neural Network}
\begin{figure}[hbt!]
  \centering
\includegraphics[scale=.4]{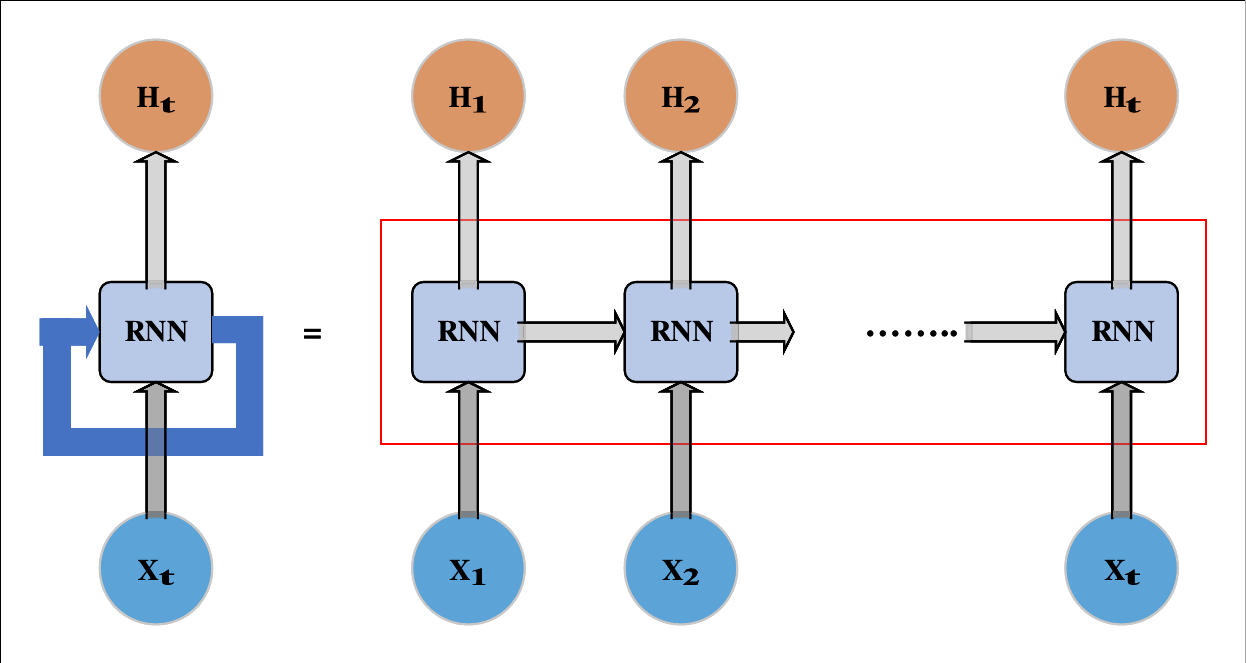}
 \caption{Basic Structure of Recurrent Neural Network \cite{viswambaran2019evolutionary,hammerla2016deep}}
\label{fig:10}
\end{figure}

Recurrent neural network (RNN) has been the center of focus in the domain of Deep Learning (DL) techniques since the 1990s. They are trained to recognize successive or time-varying patterns. A recurrent network is a neural network consisting of feedback linking \cite{fausset1994neural} which are different from other feed-forwarding neural networks \cite{svozil1997introduction}. These techniques have been applied in various kinds of problems and over the last few years, it has a wide margin of effect in the field of HAR. Depending on some feature data, the activities of humans are interpreted as time-sequential variations in numerous joint angles. For that intention, a recognizing competent in encoding time-sequential data is required \cite{park2016depth}. Fig. \ref{fig:10} shows the basic architecture of RNN. Thus, one of the consecutive DL approaches is needed for this type of work and data, that is Recurrent neural network \cite{graves2013speech}. Also, RNN models are generally better than CNN as they can construct embedded compositional portrayals of input data in the time domain \cite{uddin2019thermal}. RNNs can remember the relations between the input data and simultaneously train themselves using feedback loops \cite{viswambaran2019evolutionary}. However, basic RNNs have a vanishing gradient problem called the Long-Term Dependencies problem due to the back-spreading of an error signal over a long range of temporal intervals. This limits the processing of long-term information \cite{park2016depth,uddin2019thermal}. Searching for the solution to get rid of the vanishing gradient problem to learn the data in the protracted range, Long Short-Term Memory (LSTM) was proposed \cite{hochreiter1997long,hong2017automatic,okamoto2018recurrent}.

\par

\par
LSTM is a version of RNN makes up as input layer \cite{viswambaran2019evolutionary} and is capable of comprehending long-term reliance \cite{bakker2001reinforcement}. A block of LSTM usually contains a cell state and four different gates that are Forget gate, Input gate, Output gate, and Candidate gate . The cell state acts as an information carrier and the gates are there to control and protect that information in the cell state \cite{salem2018slim}.

The input gate $g_{ti}$ decides values that have to be updated as shown in Equation \ref{1}. The forget
gate $F_{ti}$ is used for termination of information that have to be drop out as shown in Equation \ref{2}. Cell state vector of $s_{ti}$ is used to store the long-term memory as shown in Equation \ref{3}. The output gate $o_{ti}$ decides  output as shown in Equation \ref{4} and the hidden state $h_{ti} $ can be demonstrated as a multiplication of the output gate $o_{ti} $ and the non-linear transformed cell state $s_{ti}$ as shown in Equation \ref{5}.
 \begin{equation} 
 g_{ti} = \sigma( W_{xg}x_{ti} + W_{hg}h_{ti-1} +b_g )
 \label{1}
 \end{equation}
 
 \begin{equation}
      F_{ti} = \sigma( W_{xF}x_{ti} + W_{hF}h_{ti-1} +b_F )
      \label{2}
 \end{equation}
 
 \begin{equation}
     s_{ti} = f_{ti} s_{ti-1} + i_ti\tanh ( W_{xs}x_{ti} + W_{hs}h_{ti-1} +b_s )
\label{3}
 \end{equation}
 \begin{equation}
     o_{ti} = \sigma( W_{xo}x_{ti} + W_{ho}h_{ti-1} +b_o )
 \label{4}
 \end{equation}
 
 \begin{equation}
     h_{ti} = o_{ti} \tanh(s_ti)
 \label{5}
 \end{equation}
Where $x_ti$ is an input sequence. W represents weight matrices. b represents bias vectors. $\sigma$ is the logistic sigmoid function.

\cite{murad2017deep} demonstrated 3 different methods to execute LSTM-RNN for time-series classification.

\begin{table*}[h]
\centering
	\caption{Comparison between different features of RNN in HAR.}
	\includegraphics[width=0.98\textwidth]{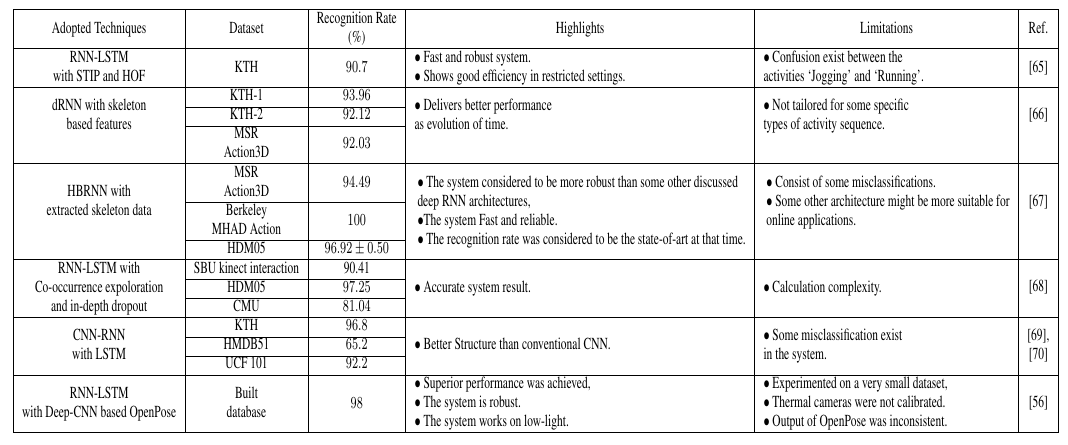}
		\label{fig:data}
\end{table*}

\begin{figure*}[hbt!]
  \centering
\includegraphics[scale=.90]{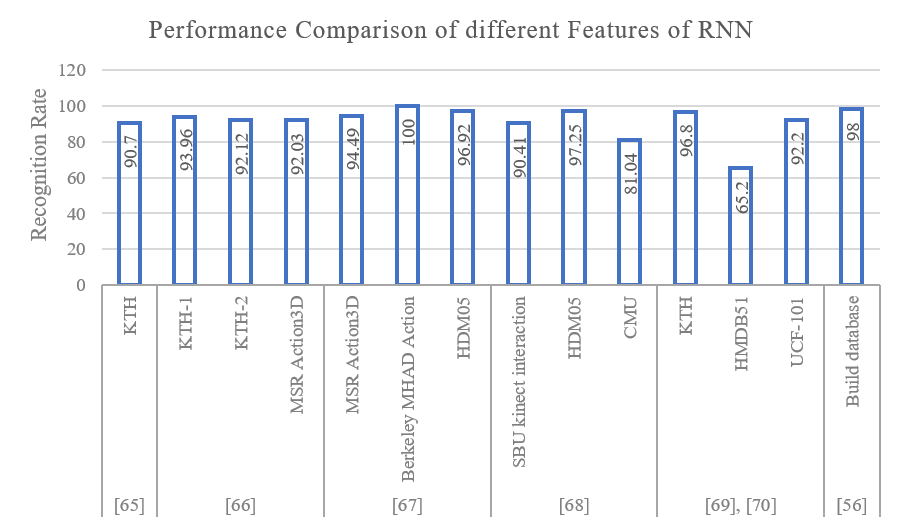}
  \caption{Histogram of the performance comparison of different features of RNN}
\label{RNNchart}
\end{figure*}

\subsubsection{Unidirectional LSTM-RNN:} 
\begin{figure}[hbt!]
  \centering
\includegraphics[scale=.35]{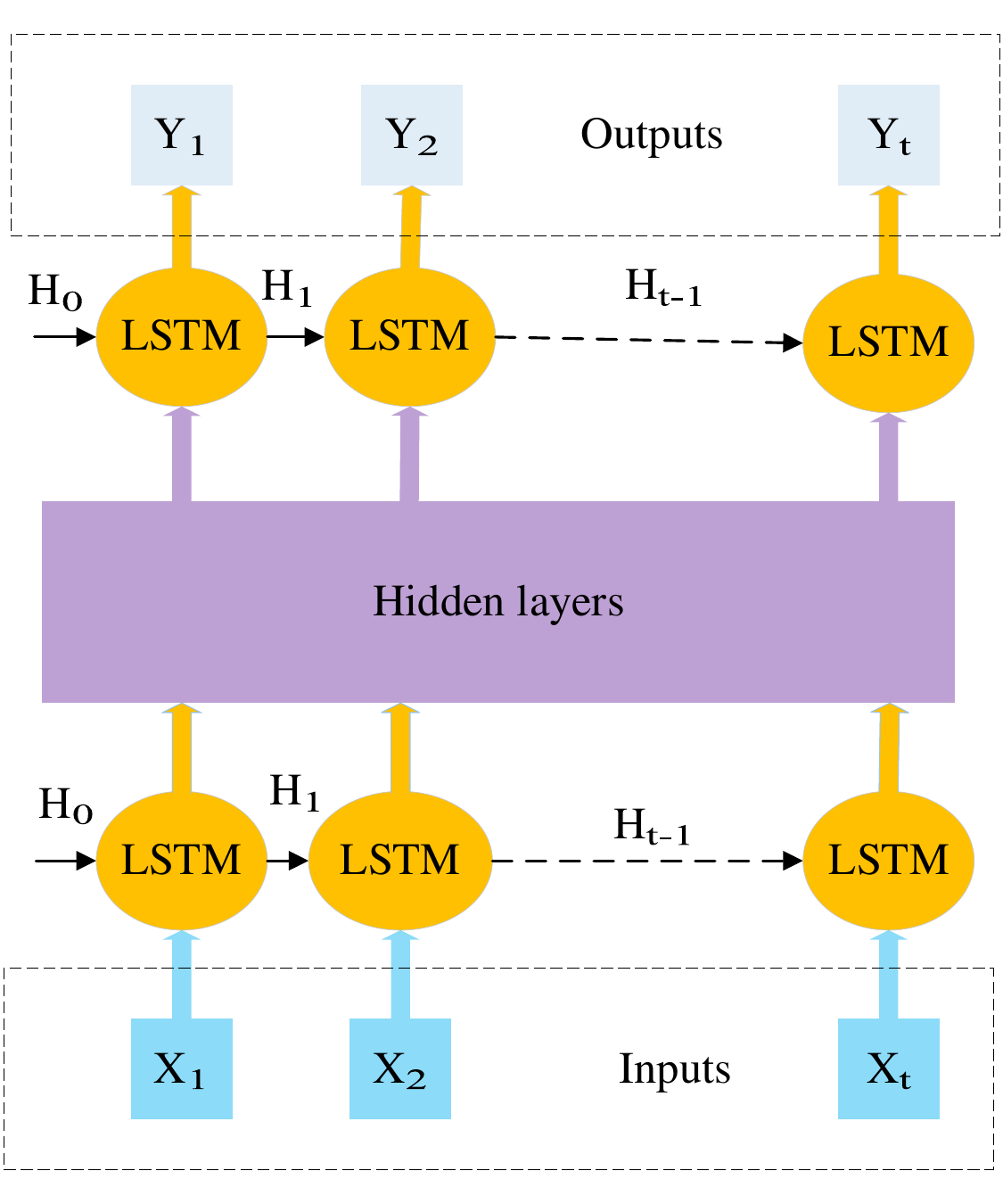}
 \caption{Unidirectional LSTM-RNN Model \cite{viswambaran2019evolutionary}}
\label{Undirectional}
\end{figure}

A series of LSTM layers are piled until the Dense layer is applied in the unidirectional RNN model. The output provided by the bottom layer is transmitted to the upper layer as data. The general premise of the unidirectional deep RNN model \cite{zen2015unidirectional} is shown in Fig. \ref{Undirectional}.

\subsubsection{Bidirectional LSTM-RNN:} 
\begin{figure}[hbt!]
  \centering
\includegraphics[scale=.43]{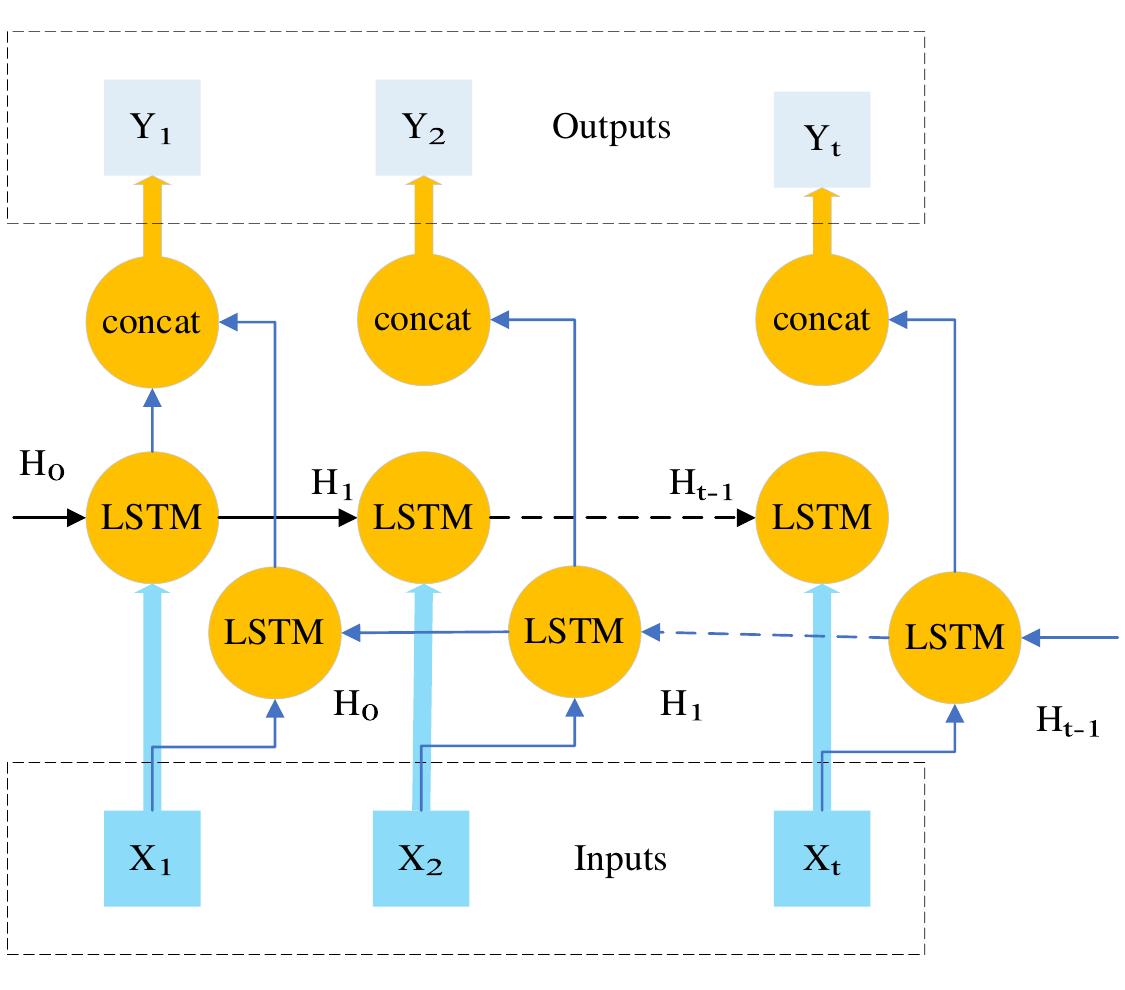}
 \caption{Bidirectional LSTM-RNN Model \cite{hammerla2016deep}}
\label{bIDIRECTIONAL}
\end{figure}

Bidirectional RNN \cite{wang2016image,li2017method} comprised of multiple tracks as portrayed in Fig. \ref{bIDIRECTIONAL}. One track for a forward shift for the upcoming state and the other for a backward shift to a previous state.
The outcome is achieved by figuring the average result within these two output tracks \cite{murad2017deep}.

\subsubsection{Cascaded LSTM-RNN:} 
\begin{figure}[hbt!]
  \centering
\includegraphics[scale=1.]{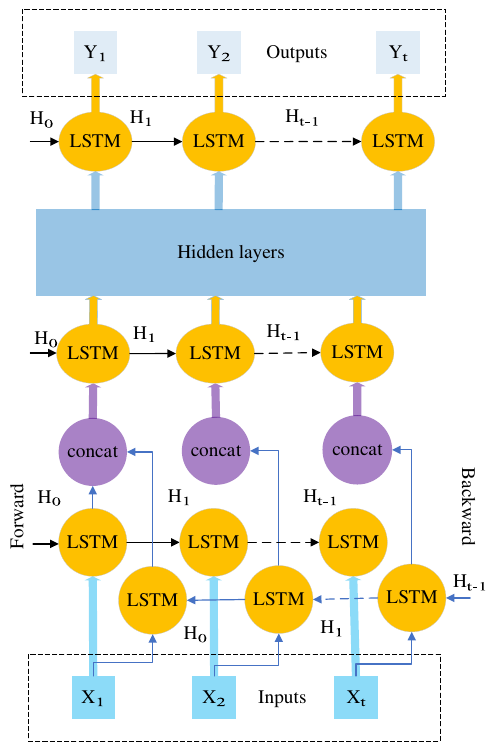}
 \caption{Cascade LSTM-RNN Model \cite{viswambaran2019evolutionary}}
\label{cascade}
\end{figure}

Both uni-directional and bi-directional layers forms together in the Cascade model. The lower one is bidirectional in nature and the remaining ones are unidirectional \cite{murad2017deep}. The general premise of the Cascade RNN model is shown in Fig. \ref{cascade}.

In the paper \cite{grushin2013robust}, the authors used LSTM-RNN with space-time interest points(STIP) and histogram of optical flow (HOF) descriptor is used for HAR in restricted settings. The authors tried to exhibit the robustness of LSTM in the real-world scenario which has an immense impact in the field of HAR-based surveillance. Harris operator \cite{harris1988combined} was extended to find STIPs, the local regions in video from which spatiotemporal features could be extracted. A HOF descriptor was computed for each of those regions where a descriptor contained 90 elements and for each of the elements there was an input neuron in LSTM. The LSTM network also included 50 memory blocks and the quantity of output neurons was equal to the quantity of activity classes which was 6 and its initial weights were arbitrarily selected. Despite having small and noisy training data and tight deadlines for making the call on activity recognition this method had shown great performance by achieving 90.7\% of the overall activity recognition rate of on KTH dataset.\par
\cite{uddin2019thermal} applied Deep-CNN based OpenPose library and non-linear discriminant feature extraction method with RNN-LSTM as the classifier for HAR. In low-light environments detecting and analyzing activities is a massive challenge as the normal colorized cameras cannot pick up the actual scenario and extracting features out of the unclear data is difficult. Therefore, the authors used a thermal camera and OpenPose library to detect human silhouettes and extract robust features. Non-linear discriminant analysis is used for further robust spatiotemporal extraction of features which are directed RNN-LSTM classifier for training the model. It had been trained by using a small database consisted of five different activities comprised of 20 long-length videos of each activity and acquired the highest overall recognition rate of $98\%$.\par
\cite{du2015hierarchical} proposed a hierarchical bidirectional RNN (HBRNN-L) model for skeleton dependent on HAR. The authors were the early ones to achieve an end-to-end solution for skeleton depended on HAR applying this method. They divided a skeleton into 5 parts where each part has a different number of joints for 3 different datasets. For MSR 3D action dataset, each subnet of bidirectional RNN has $15\times 2$, $30\times 2$, $60\times2$ and $40\times2$ neurons for the 4 layers respectively. For MHAD action \cite{ofli2013berkeley} and HDM05 \cite{muller2007documentation} datasets, each subnet of bidirectional RNN has $15\times 2$, $30\times 2$, $60\times 2$ and $60\times 2$ neurons for the 4 layers respectively. The model's efficiency was compared to five other deep RNN architectures proposed by the authors and for all three data sets, this model's performance was considered the state of the art at that time.\par
\cite{zhu2016co} also put forward an end-to-end RNN-LSTM model for skeleton dependent on HAR. The authors suggested a novel strategy of regularization learning the co-occurrence features of a skeleton as well as a new dropout algorithm for training the RNN-LSTM network efficiently. The proposed architecture contained three bidirectional layers with two feedforward fusion layers situated among them and a softmax layer in the end to give the predictions. As a consequence of massive parameter space, it is tough to learn important features despite deep LSTMs powerful learning abilities and to overcome this hurdle co-occurrence exploration was introduced. The co-occurrence of some joints of a skeleton can define human behavior inherently. The new drop-out algorithm enabled the dropping of the LSTM neuron's internal gates, cell and output response and encourages each system to learn better parameters. The authors also claimed the state of the art execution on several datasets.\par
In \cite{hammerla2016deep}, the authors introduced a DL architecture applying a bidirectional recurrent network. This introduced method executes the best performance by achieving a recognition accuracy of $92.7\%$ on the opportunity dataset. With the same dataset \cite{xi2018deep} achieved $91.3\%$ of accuracy. They used $D^2CL$ and LSTM for human activity recognition. The summary of all the applied features that are discussed in this subsection are compared in Table 3. Fig. \ref{RNNchart} Shows a histogram of comparison among different features of RNN.

\subsection{K-Nearest Neighbor}

\begin{figure}[hbt!]
  \centering
\includegraphics[scale=.55]{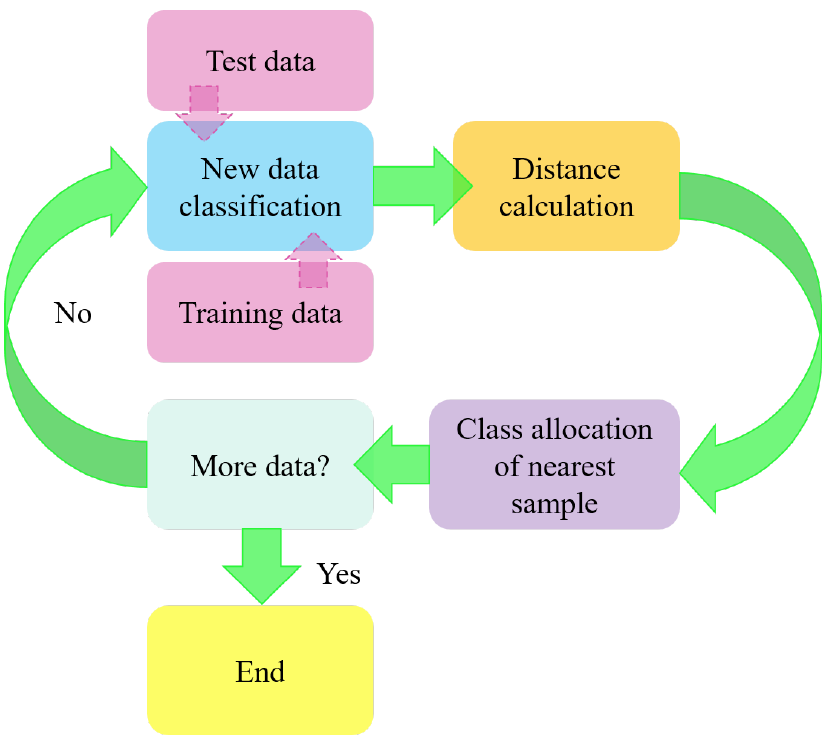}
 \caption{KNN algorithm flow diagram \cite{kaghyan2012activity,syaliman2018improving, pan2017new}}
\label{fig:5}
\end{figure}

K-Nearest Neighbor (KNN) has been serving as one well-known supervised learning algorithm. KNN is based on the instance or lazy learning because the function is solely estimated sectionally. It uses neighborhood classifications to result in the prediction of new instantaneous data skepticism. Kaghyan \cite{kaghyan2012activity} stated that three cons consist in the traditional algorithm and they are calculation complexity for simultaneous usage of all training sets, achievement depending merely on training sets, and samples consisting of zero weight difference between themselves.
 The value of k affects the prediction results \cite{kaghyan2012activity}. Fig. \ref{fig:5} shows the algorithm flow diagram of KNN. 
 
 \begin{figure}[hbt!]
  \centering
\includegraphics[scale=.32]{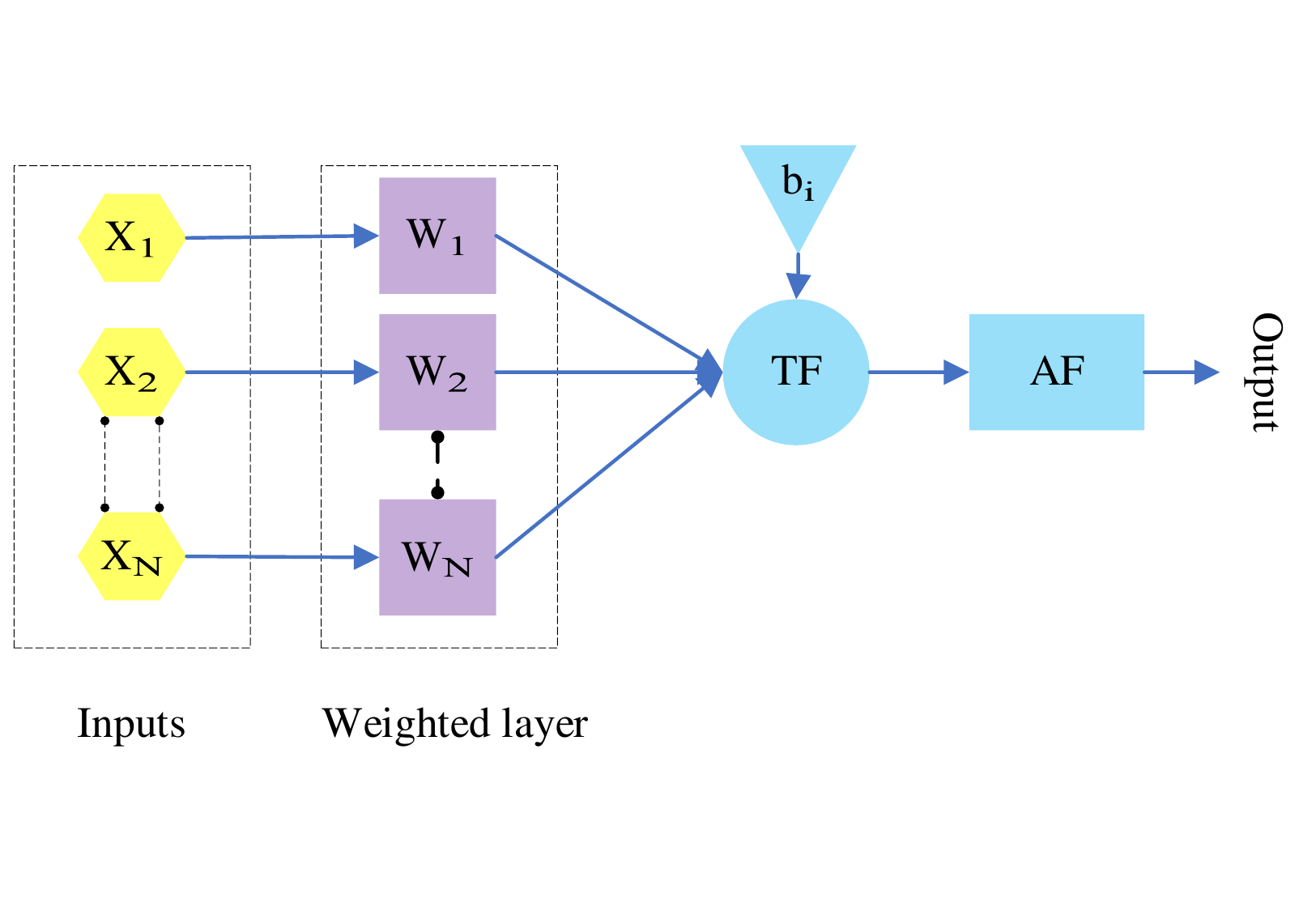}
 \caption{Basic KNN architecture \cite{syaliman2018improving}}
\label{arcknn}
\end{figure}
Everitt \cite{everitt2011miscellaneous} stated that a larger value of k filters noises while classification although makes barriers less visible among classes. And then data solely is allocated to its nearest neighbors class \cite{kaghyan2012activity}. For the optimal value determination, the bootstrap method is quite popular and is both suggested and used by \cite{suguna2010improved}.

 \par
 KNN combined with other algorithms results in greater accuracy in similar works related to recognition and detection. Fig. \ref{arcknn} shows the basic architecture of KNN. The authors of \cite{dasdetecting} used decision tables and decision trees along with KNN which lead to increased accuracy in calibrated systems for users. For continuous variables, Euclidean distance is the most common and mostly used distance metric in KNN \cite{kaghyan2012activity}. For variables that possess discrete characteristics as well as text categorization other metrics as the Hamming distance is utilized mostly \cite{norouzi2012hamming}.
 The authors of \cite{kaghyan2012activity} stated the frequent test set that nominates to rule over the prediction vectors over new ones as a major drawback. These sets tend to come up while the neighbors are being computed for their large amount of data. It happens when the distribution class is being pitched. To overcome this drawback the classification is weighted that of its distance to the k nearest neighbor.
 \par
 The authors of \cite{mitani2006local} claimed LMKNN to be proven to enhance classification performance with simple and effective nonparametric classification techniques. It is also said to lessen the consequence of the anomalies which exist, particularly for small-sized data \cite{pan2017new}. This process could be represented by equations as follows \cite{syaliman2018improving}: 
 \begin{equation}
     D(l,z)=||l-z||^2=\sqrt{\sum_{j=1}^{N} |l-z|^2}
 \end{equation}
 
 The goal is to calculate and find the following equation \cite{pan2017new}:
 
 \begin{equation}
     m_w^k= \frac{i}{k}{\sum_{i=1}^{k} l_{i,j}^{N N}}
 \end{equation}
 For LMKNN the quantity of nearest neighbors from all class is noted as the value of k in training data \cite{pan2017new,syaliman2018improving}. 
\par

In \cite{danades2016comparison}, the authors computed the accuracy of KNN and SVM to achieve a mean accuracy between SVM and KNN which is obtained $92.40\%$ and $71.28\%$ respectively.
In \cite{tamatjita2016comparison} the correlation is shown between kNN and NCC in which the research result unveiled that NCC obtains an accuracy margin between $96.67\%$ and $33.33\%$, whereas kNN only produced an accuracy margin between $26.7\%$ and $22.5\%$. Their research unquestionably showed the poor accuracy of kNN compared to NCC.
The authors of \cite{brown2017predicting} noticed that KNN produces the best accuracy of $48.78\%$ when the value of k is k=$8$.

\begin{table*}[h]
\centering
	\caption{Comparison between different features of KNN in HAR.}
	\includegraphics[width=0.98\textwidth]{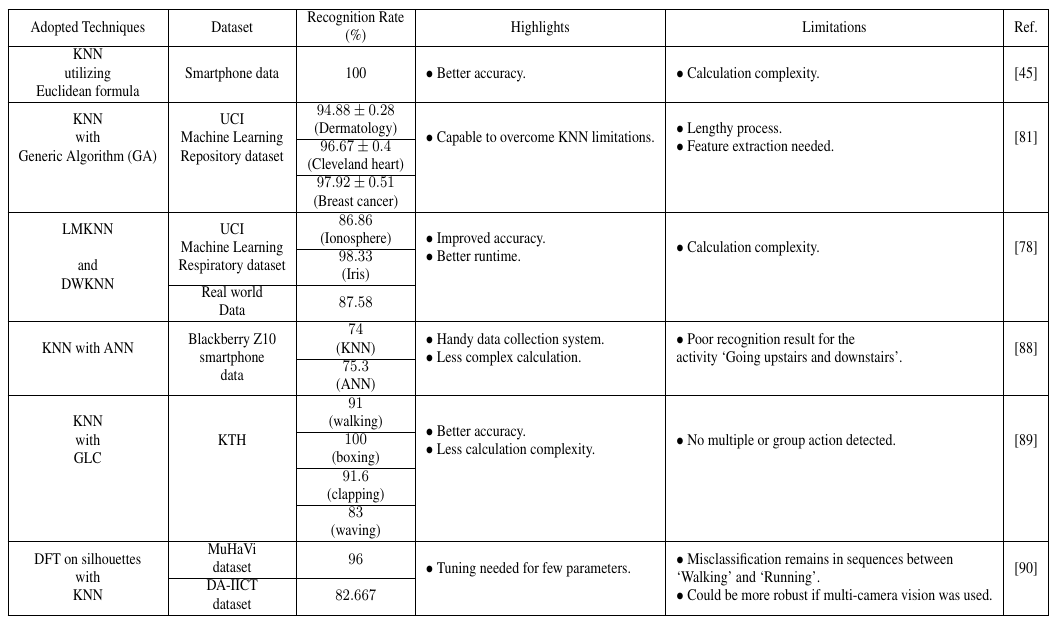}
		\label{fig:data}
\end{table*}

\begin{figure*}[hbt!]
  \centering
\includegraphics[scale=.90]{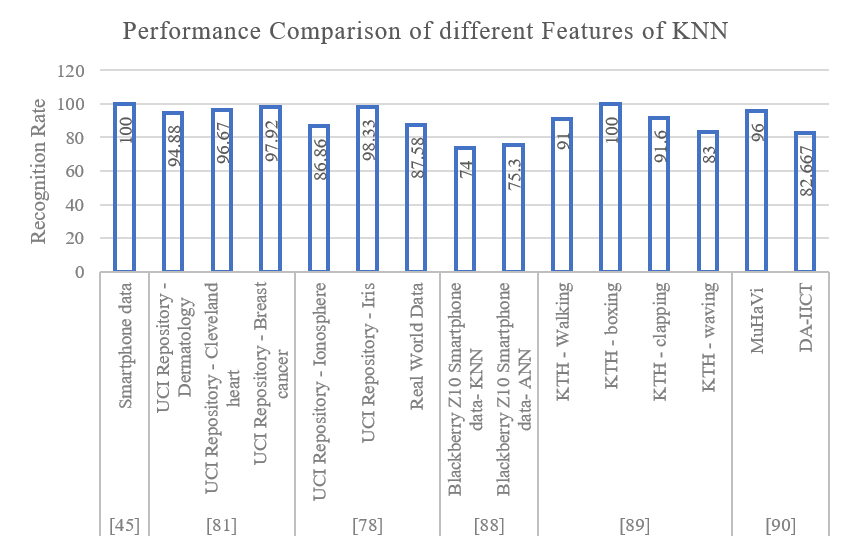}
  \caption{Histogram of the performance comparison of different features of KNN}
\label{KNNchart}
\end{figure*}

\par
In \cite{kaghyan2012activity}, the Euclidean formula is used along with KNN to recognize various human activities from the data gained from the Tri-axial accelerometer. KNN is used for analyzing the raw data gained from the accelerometer. To separate and calculate 3-dimensional attributes Euclidean formula is applied. This is because KNN works with multidimensional data and standard classifiers are unable to classify raw data. The system was optimized considering k=1. This is because the device, smartphone, is not optimal to compute complex calculations especially when the calculation becomes difficult for k being number(s) higher than $1$.
\par
\cite{suguna2010improved} overcame the classic KNN limitations regarding HAR such as calculation complexity, majority voting system, training set dependent performance, weight balancing between sample data. They proposed Genetic KNN (GKNN) which is a fusion of Genetic Algorithm (GA) with KNN. Unlike classic KNN, GKNN uses GA to directly employ the K-neighbours and after taking the k-neighbor, it classifies distance measurement between test data. This helps to avoid the consideration of similarities and weight categorization between all the test data for each iteration which simplifies the calculation.

\par
The authors of \cite{syaliman2018improving} used LMKNN and DWKNN instead of classic KNN for accuracy improvement of activity recognition. Classic KNN deploys a vote majority system which is a reason for KNN having low accuracy. Vote majority system passes over the proximity between test data \cite{gou2011novel}. For this reason, distance differs greatly from each of the nearest neighbors against each of the test data. To avoid this phenomenon they used LMKNN and DWKNN instead of classic KNN. Using LMKNN the closest distance was calculated and sorted from the training data for a fixed value of k and DWKNN is used for computing and picking out a class containing the highest average of weight for the latest class assigned for the test dataset. They found an increase of $5.16\%$ in average accuracy. Though acknowledged as simple, instinctive, and easy to apply, the kNN algorithm is observed to produce lower accuracy than other classification algorithms, particularly when compared \cite{syaliman2018improving}.
\par
\cite{sang2015human} recognized daily human activity employing just the data obtained from an accelerometer and the gyroscope of a smartphone. They used Artificial Neural Networks (ANN) along with KNN. They considered that the smartphone would be located in the user's (human) pocket. The data or signal from both accelerometer and gyroscope is preprocessed and extracted to their prime features by using different methods. Then the extracted features were classified by KNN using the Euclidean formula. This is to compute the distance between different sample data in test data. After classification ANN is used to recognize the user's (human) activity. This helps to get precise results on different activities. They successfully pulled out on average $76\%$ to $100\%$ overall accuracy on different activity recognition.
\par
\cite{kumari2011human} focused on data preprocessing and they used different methods for operating background subtraction. They used Gaussian Mixture Model (GMM) for background subtraction. This helps reducing data noise at a great level. Discrete Fourier transform (DFT) is used to extract features of the test data. This is because the Fourier domain has multiple advantages such as having a greater range in imagery, being able to calculate and store data in a float value, ensuring featured data being preserved, having important data from original samples. After the feature is extracted with DFT, classical KNN is used to classify features and recognize human action. They used the classical KNN algorithm to make the calculation simple and robust. The summary of all the techniques applied in this sub-section is compared along with their advantages \& limitations in Table 4. Fig. \ref{KNNchart} Shows a histogram of comparison among different features of KNN.
\par

\subsection{Support Vector Machine}

The Support Vector Machine (SVM) happens to be a supervised learning model consisting of work-associated learning algorithms. SVM represents sort of a statistical learning method where several principles are inaugurated to obtain classifiers having accurate results \cite{vapnik2013nature}. For Data separation it builds many hyperplanes in infinite-dimensional space. Both linear and nonlinear operations can be acquired by SVM. For linear separation, margin operation is applied, for obtaining the maximum distance or deviation from the data to the margin. This is the hard margin method. It is only applicable when the data are linearly separated. when not, a loss function is added in the function classifier although it is bound with a tolerance factor \cite{hasib2021comprehensive}. This is called the soft margin method for linear operation.

\begin{figure*}[hbt!]
  \centering
\includegraphics[scale=.45]{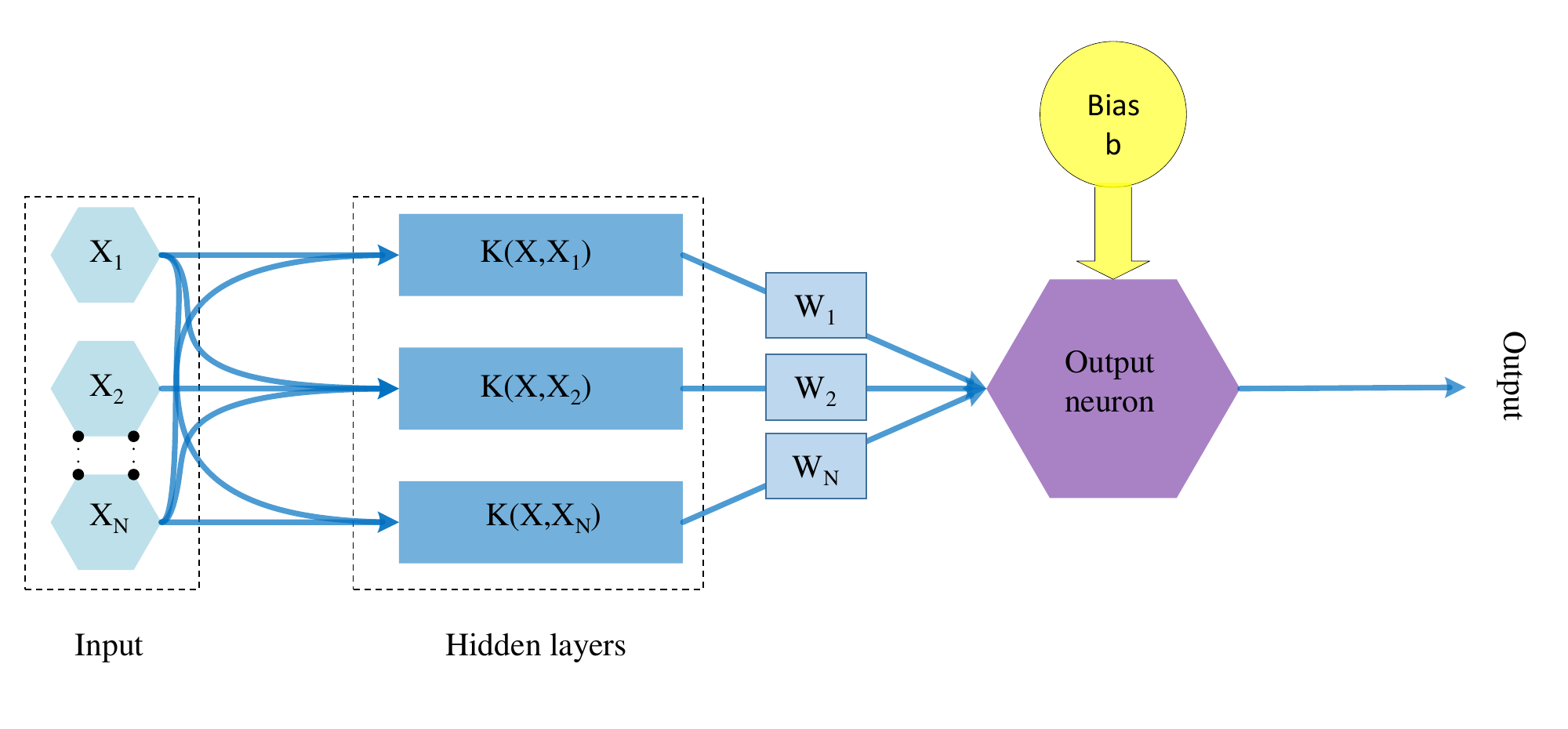}
 \caption{Basic SVM Architecture \cite{chathuramali2012faster,althloothi2014human}}
\label{arcsvm}
\end{figure*}

SVM is mainly and broadly used for data classification. Fig. \ref{arcsvm} shows the basic architecture of SVM. Classification is executed by using hyperplanes in the featured space by making several linear decisions \cite{chathuramali2012faster}. It has several extensions as Support vector clustering \cite{ben2001support}, multi-class SVM \cite{lee2004multicategory}, kernel PCA, density estimation and regression \cite{chathuramali2012faster}, and others.
\par
\cite{althloothi2014human} demonstrated a method that portrays both 3D joint motion features and depth contours. This method is to represent, recognize, and classify several human activities using multi SVM. For very large training data sets of HAR, the scale of SVM runs very accurately and is also very much cost-effective but the complexity rises incrementally with the increment of numbers of any dataset although classification operation in SVM is served independent  \cite{chathuramali2012faster}. However, the authors of  \cite{muller2001introduction} stated that hyperplane separation towards linear classifiers can be put down as follows:

\begin{equation}
v= sgn((s \times U)+c) 
\end{equation}

And the classification condition for separating hyper plane classifier worth zero training loss is:

\begin{equation}
  v={((s \times U)+c)\geq 1}; i=1,.......,n
\end{equation}
\par 
Where 's' expressed as normal vector of hyper-plane, 'v' expressed as the label of training pattern, 'U' expressed as input space, and 'i' expressed as a counter of patterns \cite{muller2001introduction}.
\par
This learning method intends to express the optimal hyperplane. The goal is to minimize \cite{muller2001introduction}:

\begin{equation}
  \frac{1}{2}||s||^2
\end{equation}
\par

  

\par

\begin{table*}[h]
\centering
	\caption{Comparison between different features of SVM in HAR.}
	\includegraphics[width=0.98\textwidth]{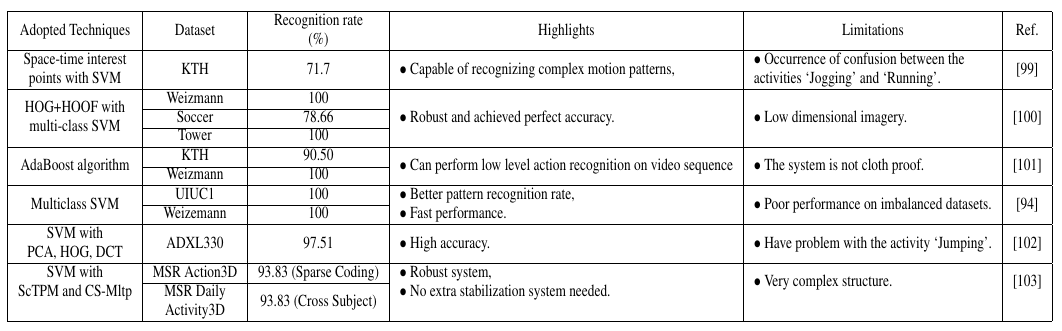}
		\label{fig:data}
\end{table*}

\begin{figure*}[hbt!]
  \centering
\includegraphics[scale=.90]{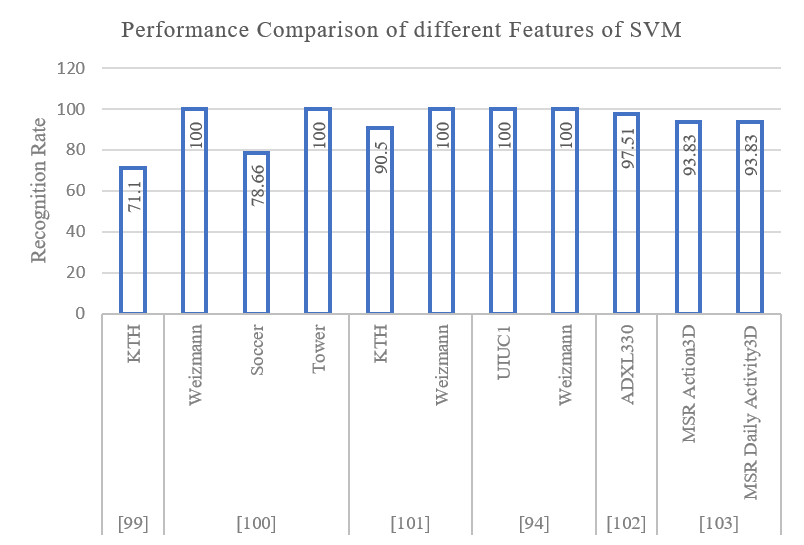}
  \caption{Histogram of the performance comparison of different features of SVM}
\label{svmchart}
\end{figure*}

\cite{escorcia2013spatio} applied the Spatio-temporal model along with human-object interaction for HAR in videos. The spatial and temporal model is evaluated to identify the human-object relationship. This model is dependent on steams. The acquired targets, as well as the humans, are spotted and the positions for both humans and objects that are in the streams are encoded with a successful explanation. Human-object features are extracted using the detection window. SVM classifier is used to classify to describe the estimation of sequences of intervals among the entire video. They get an accuracy of $96.3\%$ and $98\%$ which is been evaluated in various checks on various data- sets. Although having a remarkable result, the system failed to consider the location poses and objects surroundings in the process.
\par
\cite{schuldt2004recognizing} employed local space-time features along with SVM for the enhancement of results in complex human activity recognition. They created a new video dataset for this purpose which contains $2391$ sequences with $25$ subjects. Their work aims to show that Local Features (LF), that is, regarding Spatio-temporal interest points the local measurements conceivably pull off recognizing complex human activities along with SVM classifier. They achieved a recognition rate of $71.7\%$ recognizing complex human activity using LF with SVM. SVM is used because it works well space-time margin assigning an optimal hyper-plane measurement.
\par
In \cite{chathuramali2012faster}, the authors used multiclass SVM to recognize human action. The motivation behind this is to overcome the limitations of classic classifiers, such as longer training time, a larger size of the feature, limited space-time margin. They used LF as well as motion descriptors for feature extraction. Background subtraction is used for the extraction of profiles of the character samples. It helps to pull off the activity recognition task in both high and low-resolution images. Multiclass SVM is used as a classifier to reduce computational and training time though it lacks accuracy at the larger dataset. They successfully gained a result of which the recognition rate was $100\%$ with UIUCI and Weizmann datasets.

\par
\cite{chen2009recognizing} utilized Histogram of Oriented Gradients (HOG) \cite{ahamed2018hog} and Histogram of Oriented Optical Flow (HOOF) alongside with SVM in Activity recognition. To describe human pose HOG is applied. It mainly work as a human indicator \cite{dalal2005histograms} but it is robust enough to successfully apply in HAR problems \cite{hatun2008pose}, \cite{lu2006simultaneous}, \cite{thurau2007behavior}. But only human detection is not enough for Activity recognition. That is why HOOF is implied as it can characterize and detect human movements by extracting feature vectors with the help of optic flow. Features are extracted with dimensional space-time reduction for every type of featured vector. Multi-class SVM was deployed to classify human action and for computing nonlinear relation between actions, radial basis function (RBF) kernel is used. For the assessment of the best classifier in a particular dataset, they used the grid search technique.
\par

\par
In \cite{fathi2008action}, the authors approached in a different way towards HAR. Instead of just assigning LFs, they moved on to figure-centric motion representations of human actions. This representation is gained by employing a detector or motion tracker at the very beginning of input image sequencing.
They used three layers for training and classifier learning so that they can distinguish between different activities. AdaBoost is used to extract pieces of information from both low and mid-level image sequences. AdaBoost is run for a second time to merge different parts of different image sequences. For classification, they used a one versus all approach to differentiate classes and recognize different activities using AdaBoost and Multi-class classifier with Hamming decoding. With this approach, they have successfully achieved an efficient and fast runtime compared to others with a range of about $0.2$ to $4$ seconds per frame. \cite{luo2014spatio} introduced a framework that is quite different for HAR. Depend on RGB and depth features, the framework serves to accompany SVM.
\par
\par
\cite{he2009activity} Utilized SVM along with Discrete Cosine Transform (DCT) as well as PCA with data from a tri-axial accelerometer for HAR. Features are extracted with the help of DCT to carry out the most effective ones. Then feature reduction is employed to lessen data proportion of as well as from original data. PCA is used to do this reduction operation. After reduction SVM is employed for the classification of the data and human action detection. They used SVM because the training dataset was limited. To overcome this problem they used the leave-one-subject-out validation \cite{wang2005human} approach. And they achieved a rich recognition rate of $97.51\%$ in different types of daily human activity. A comparison of all these previously applied techniques of SVM is shown in Table 5. Fig. \ref{svmchart} Shows a histogram of comparison among different features of SVM.
\par

\section{Abnormality Detection using Unsupervised Learning}

While supervised learning algorithms perform well in some cases, there exist several challenges which have to be addressed to get good performance from these models. Supervised learning-based methods work based on ground truth or data with labels that have to be provided by a human expert. To get these algorithms to perform well, the labeled data has to be a hundred percent accurate and the dataset has to be large. Labeling a large amount of data is no easy task for any human expert, and is often, quite impractical, and not a viable solution. Also, in many cases, a human expert may fail to find the underlying structure of the data which can be vital to detect certain behaviors. For these reasons, Unsupervised learning-based methods were introduced into the HAR problem.

Unsupervised Learning algorithms possess a vital impression in HAR for anomaly detection.  Unlike supervised learning, these methods learn untagged data, which saves the trouble of labeling thousands of data. This method does not require any past knowledge or in simple words, this method is not pre-trained for any operation. These methods find identical patterns in the data, group them accordingly, saving the trouble of hand-labeling. Unsupervised learning-based methods find the underlying structure of the data, which may be very difficult sometimes to detect and label properly. The idea of this method is that the normal events happen frequently but abnormal event seldom happens from which the abnormality in daily life events can be easily detected.

\subsection{K-Means Clustering}
The k-means is a popular unsupervised clustering algorithm. It is also recognized to merge the local minimum value of the Euclidean distance.  K-means algorithm applies the Sum of Squared Error (SSE) and Total Sum of Squares (SST) to compute the status of the cluster. In a k-means algorithm, the lower the SSE or SST value, the better the clustering. SSE and SST can be mathematically represented by equation \ref{sse} and equation \ref{sst} respectively where  x is a vector representation of an object, $C_i$ represents the $i^{th}$ cluster, $c_i$ is the centroid of $C_i$, and $c_m$ represents the mean of all the allocated points \cite{ahmed2016survey}. 
\begin{equation}
\textrm{SSE}= \sum_{i=1}^{k}\sum_{c_{i}}^{}dist(c_{i}, x)^{2} \quad \textrm{and} \quad\forall_{x}\in C_{i}
\label{sse}
\end{equation}
\begin{equation}
    \textrm{SST}= \sum dist(c_{m}, x)^{2} \quad \textrm{and} \quad\forall_{x}  \; \textrm{in dataset}\; D
    \label{sst}
\end{equation}

\begin{table*}[h]
\centering
	\caption{Comparison between different features of Unsupervised learning in HAR.}
	\includegraphics[width=0.98\textwidth]{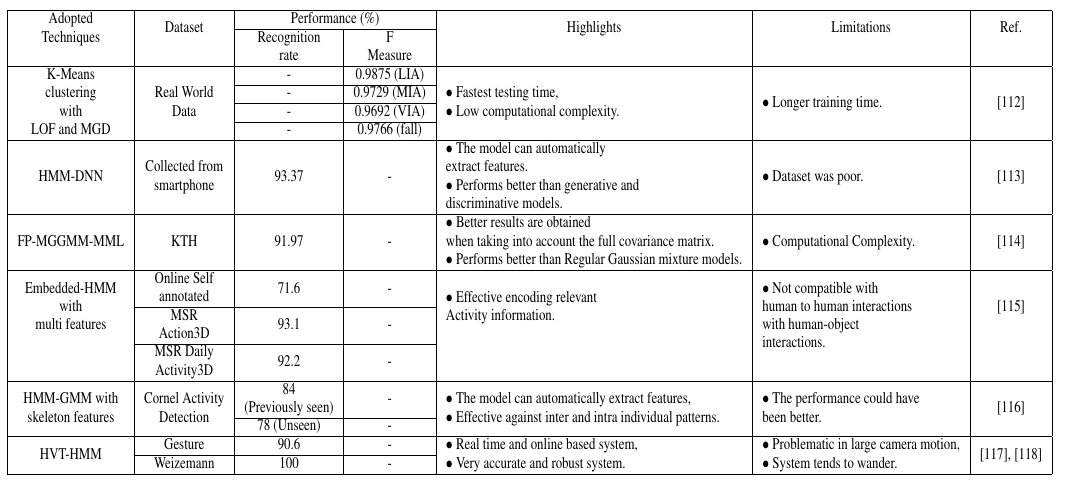}
		\label{fig:data}
\end{table*}

\begin{figure*}[hbt!]
  \centering
\includegraphics[scale=.90]{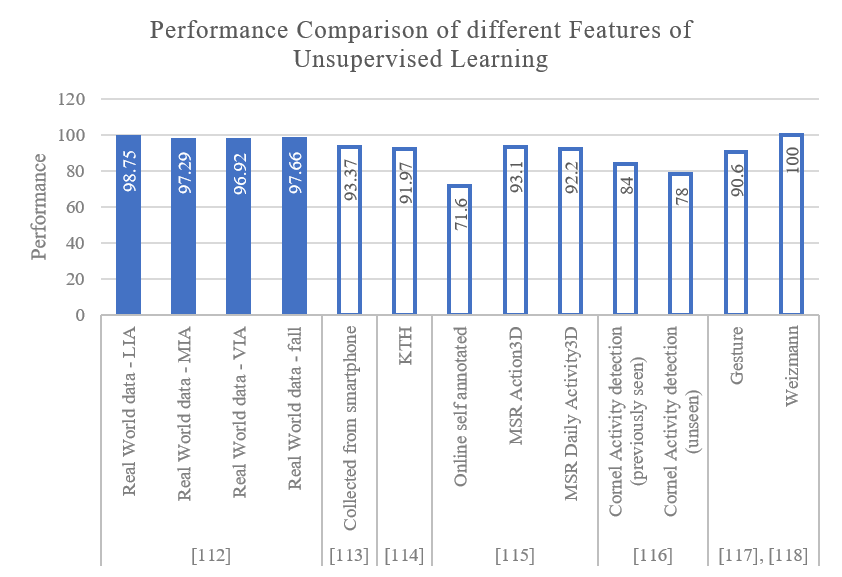}
  \caption{Histogram of the performance comparison of different features of Unsupervised Learning}
\label{unchart}
\end{figure*}


\par
 In \cite{zhao2018user}, the authors put forward an adjustable technique formed on K-Means clustering, local outlier factor (LOF) as well as multivariate Gaussian distribution (MGD) for HAR. They personalized their classifier to recognize human activity into four levels that are light-intensity activity (LIA), moderate-intensity activity (MIA), vigorous-intensity activity (VIA), and fall. Their proposed technique copes with user-specified data to learn and detect human activities. Their essence is to replace the initial centroids of the unclustered data with the k centroids of the activity dataset that is collected before. They achieved the fastest testing time of $0.02$ seconds and an F-measure of $0.9875$, $0.9729$, $0.9692$, and $0.9766$ for LIA, MIA, VIA, and fall respectively.
 In \cite{chen2015detection} introduced a novel K-means clustering method for abnormality detection in a crowded area based on acceleration feature. The acceleration feature is based on Newton's 2nd law and it utilizes the assumption that when an abnormal event is taking place the acceleration of the movements of people in the crowd would increase. The frames of the videos are represented using velocity and acceleration fields and are divided into patches where K-means clustering was used to detect the abnormal situation adopting acceleration features. The proposed model was used on UMN and PETS2009 datasets and showed better performance than compared methods.

\subsection{Hidden Markov Model}
\begin{figure*}[hbt!]
  \centering
\includegraphics[scale=.55]{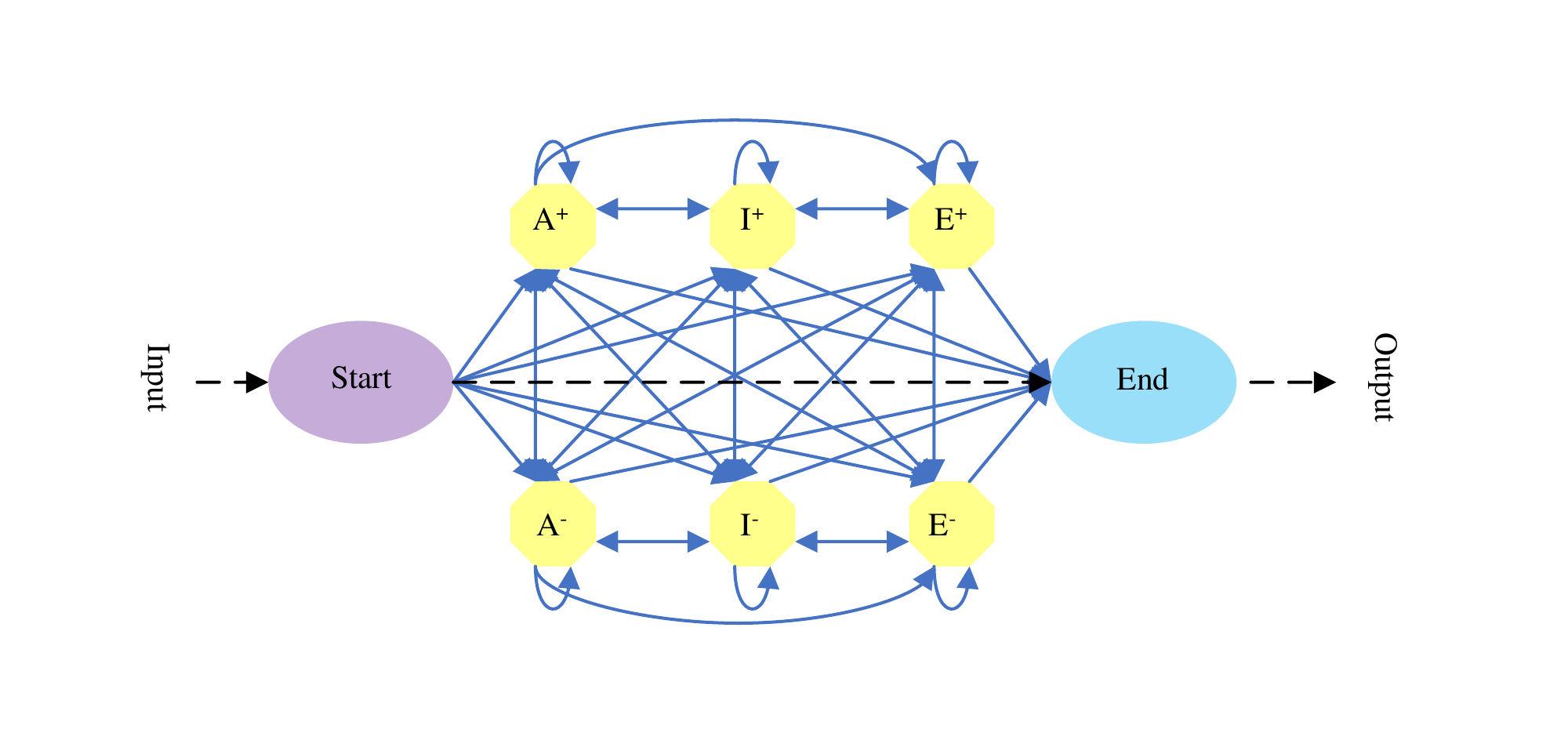}
 \caption{Basic HMM Architecture \cite{zhang2015human}}
\label{archmm}
\end{figure*}

Hidden Markov Model (HMM) emerges from a basic chain of Markov. It contains observation vectors that are used for data item sequencing. The states of the system are obtained through this statistical model \cite{zhu2016high}. Fig. \ref{archmm} shows a basic architecture of HMM. It is an important way to automatically recognize speeches and processing signals \cite{bilmes2002hmms}. Interaction among individuals is estimated based on possibility distribution via HMM with sequence learning dispensation \cite{kulkarni2020survey}. As far as recognition techniques are concerned, HMMs are extensively used in numerous HAR systems so far, as they are competent in time-varying pattern decoding \cite{piyathilaka2013gaussian, iengo2014continuous, jalal2013human}. Many types of HMM models are being used for HAR such as continuous HMM, probability-based graphical HMM, HMM-DNN, HMM-GMM models are seen to be used in some of the reviewed literature.

\par
\begin{figure}[hbt!]
  \centering
\includegraphics[scale=.85]{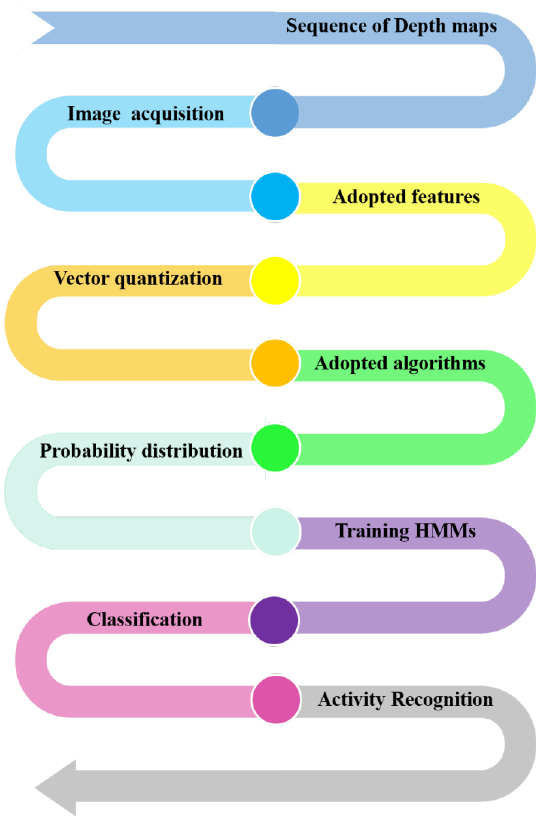}
 \caption{Simple Flow Diagram of HMM Algorithm \cite{jalal2017depth}}
\label{flowhmm}
\end{figure}

\cite{jalal2017depth} proposed a novel approach of embedded HMM with multi-features using depth video sensor for HAR. The authors introduced a real-time body parts tracking system for extracting human silhouettes from noisy backgrounds and several symbolic features counting in-variants, multi-view differentiation, and Spatio-temporal body joints, which have been combined to explore change in gradient orientation and motion of specific body parts. To recognize human activities these sub-features were unified together forming a multi-feature vector that sizes $689\times1$ and trained in the proposed embedded HMM where active feature regions of human body joints were specifically focused on. The authors constructed a new online dataset regarding human activity to evaluate the proposed model along with three other existing datasets and found a commendable recognition rate atop the state-of-the-art performance at that moment. Fig. \ref{flowhmm} shows a simplified flow diagram of the HMM classifier. The authors of \cite{jalal2015individual} introduced a human activity recognition system employing depth silhouettes context features using advanced HMMs applying depth sensors that provide higher performance on recognition rates over the state-of-the-art methods. The recognition rates are $57.69\%$ on IM Daily Depth Activity dataset and $83.92\%$ on the MSR Action3D dataset.
In \cite{najar2019unsupervised} the authors proposed unsupervised learning of finite multivariate generalized Gaussian mixture model. The authors have developed a novel learning algorithm based on a Fixed-point covariance matrix estimator combined with the Expectation-Maximization (EM) algorithm and to overcome the model selection problem and an appropriate minimum message length (MML) criterion was proposed. The proposed model has shown promising outputs in the KTH dataset. 
\par
\cite{piyathilaka2013gaussian} employed a Gaussian Mixture Model (GMM) depended HMM using 3-D skeleton features for HAR provided by RGB-D camera. Being a set of multinomial Gaussian distributions, Gaussian mixtures can cluster data into various groups therefore, each of the body positions conceivably is formulated as a multitude of the multinomial distribution. The authors employed the following model applying Byasian Network ToolBox found in Matlab. The mixture with Gaussian parameters was embarked applying the K-mean algorithm and optimized by expectation maximization (EM) algorithm.
\par

However, \cite{zhang2015human} acclaimed Deep Neural Networks (DNNs) to work better in modeling the emission distribution of HMM than generative Gaussian Mixture Models (GMM) or discriminative Random Forest(RF). The authors recommended HMM-DNN methods over HMM-GMM and HMM-RF methods as they have the capability to automatically extract important features that could be lost or time-consuming in the case of HMM-GMM and HMM-RF. The parameters such as weights and biases of the proposed HMM-DNN model were randomly initialized and optimized using a back-propagation algorithm and it was trained to apply stochastic gradient descent with a mini-batch size of 100 training samples where 1000 epochs at a rate of learning of 0.1 with a weight decay of about 0.0002 were run. The experimental results found the HMM-DNN process generates an accuracy of $93.37\%$ when HMM-RF generates $85.69\%$, and HMM-GMM generates $84.88\%$ of accuracies sequentially.\par 

A comparison of these unsupervised learning techniques applied to solving the HAR problem is presented in Table 6. Fig. \ref{CNNchart} Shows a histogram of comparison among different features of unsupervised learning techniques where the filled pillars represent f1 score $\times 100\%$ and the hollow pillars represent recognition rate.

\section{Datasets of HAR}
The importance of datasets is increasing drastically as the approach to the solution of the HAR problem is gradually favoring the data-driven deep learning models because of their massive advantages like accurate results, adaptability, and quick decision-making ability. For smoothing the path to clearly distinguish between the abilities of these learning models various datasets consisting of labeled human activity have been created. To cope with the continuously advancing field of HAR the evolution of video-based datasets has also come a long way. In this section, some of the most used datasets in the research of HAR are discussed.

\begin{table*}[h]
\centering
	\caption{Overview of Some Popular Video Datasets Used in HAR.}
	\includegraphics[width=0.68\textwidth]{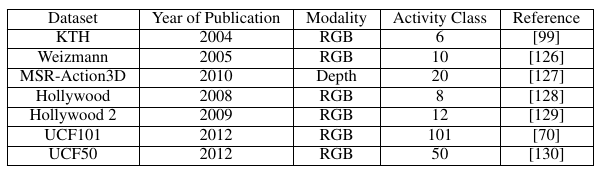}
		\label{fig:data}
\end{table*}

\subsection{UCF50}
The UCF50 \cite{reddy2013recognizing} is an activity recognition dataset holding fifty (50) action Classes. It consists of 6676 practical videos attained from YouTube. This dataset is an enlargement of the dataset UCF11. The UCF11 contains eleven (11) action classes. Many of the available activity recognition datasets are impractical as well as staged by actors. In which many of the activities are not executed well. In UCF50, the principal focus is to provide the computer vision world with a dataset possessing true-to-life videos. The dataset is very intricate because of having large fluctuations of in-camera motion, appearance, and pose of the central object, object computation, focus, cluttered environment, lighting limitations, and so on. The videos are arranged into 25 groups of the entire dataset, and each group has more than 4 action clips. The video clips share some common features in a group as it is done by the same actor in the same environment.

\subsection{UCF101}
The UCF101 \cite{soomro2012ucf101} is an activity recognition dataset. It consists of practical videos gathered from YouTube. It holds 101 action classes of videos. This dataset is an augmentation of the UCF50 dataset which has fifty (50) action classes. Possessing 13320 video clips from 101 action classes, UCF101 affords the largest multifariousness in terms of actions and with the presence of large inconstancies in-camera motion, appearance, and posture of the central object, the balance of the object, location, cluttered environment, lighting states, and so on. It possesses the unique reputation of being the most challenging data set to date. UCF101 intends to promote moreover research toward activity recognition by learning and investigating new lifelike action classes. 

\subsection{Weizmann}
Weizmann actions dataset \cite{blank2005actions} consists of 10 different types of natural activities. There are 90 videos in the dataset. The videos are of a low resolution of 180x144 and 50fps. Testing is executed in a leave-one-out manner on a per-person basis. Training is done on eight subjects and testing on the remaining subject and all its videos. An adjusted camera and a simple background were adopted. Then the video clips of the dataset was recorded . Most of the activities were performed by a single actor.

\subsection{KTH}
The KTH \cite{schuldt2004recognizing} video database containing 6 types of human activities. It is one of the most cited datasets. It was conducted numerous times by 25 subjects in four separate scenarios. The database consists of 2391 sequences. All sequences were taken over uniform backgrounds. A static camera was used to record. The sequences were of 25fps frame rate. The continuities were downsampled to the spatial resolution of 160x120 pixels. And it consists of a length of four seconds on average. There lie 25x6x4=600 video clips for each succession of 25 subjects, 6 actions, and 4 scenarios. It has got high inter-class variation which has to lead it to its success.

\subsection{Hollywood}
The Hollywood \cite{laptev2008learning} dataset, also called HOHA, was created at IRISA institute, France. The datasets associated with HAR are different. They are explained by automatic script-to-video order along with text-based script classification in movie scenarios. Hollywood has 8 activity classes collected from 32 movies. There consists dynamic background as the video scenes were taken from different movies and it makes a tough task for any HAR model for activity recognition. Of these video scenes, 20 make the train set and 12 the test set. The train is separated into two sets, an automatic annotation set possessing 233  video clips and a manually verified set possessing 219 video clips. The test set has 211 video clips that are manually validated. This dataset is very much challenging as the camera viewpoints change simultaneously.

\subsection{Hollywood 2}
The Hollywood 2 \cite{marszalek2009actions} is an extension of Hollywood and has a total number of 12 categories. Four new additional activity catagories are added and they are driving a car, eat, fight, and run. The dataset aims to provide a complete benchmark for human activity recognition in realistic and challenging environments. A total number of 69 movies used in the data set are split into 33 trains and 36 test clips. Scene video samples are produced next to employ script-to-video alignment. Labels of test scene samples are verified manually. 

\subsection{MSR-Action3D}
The MSR-Action3D \cite{li2010action} dataset is an activity dataset of depth sequences created by Wanqing Li at Microsoft Research Redmond. It is frequently used to benchmark different depth based HAR approaches. The video scene was captured by a depth camera. This dataset contains 20 action types with 10 subjects. Each of the subjects in this dataset performs each activity more than once. There are 567 depth map sequences of 640x240 resolution in total. The video data was recorded by a depth sensor comparable to the Kinect device.
A summary of all the discussed dataset is presented in Table 7.

\section{Challenges and Future Trends}
\begin{figure}[hbt!]
  \centering
\includegraphics[scale=.4]{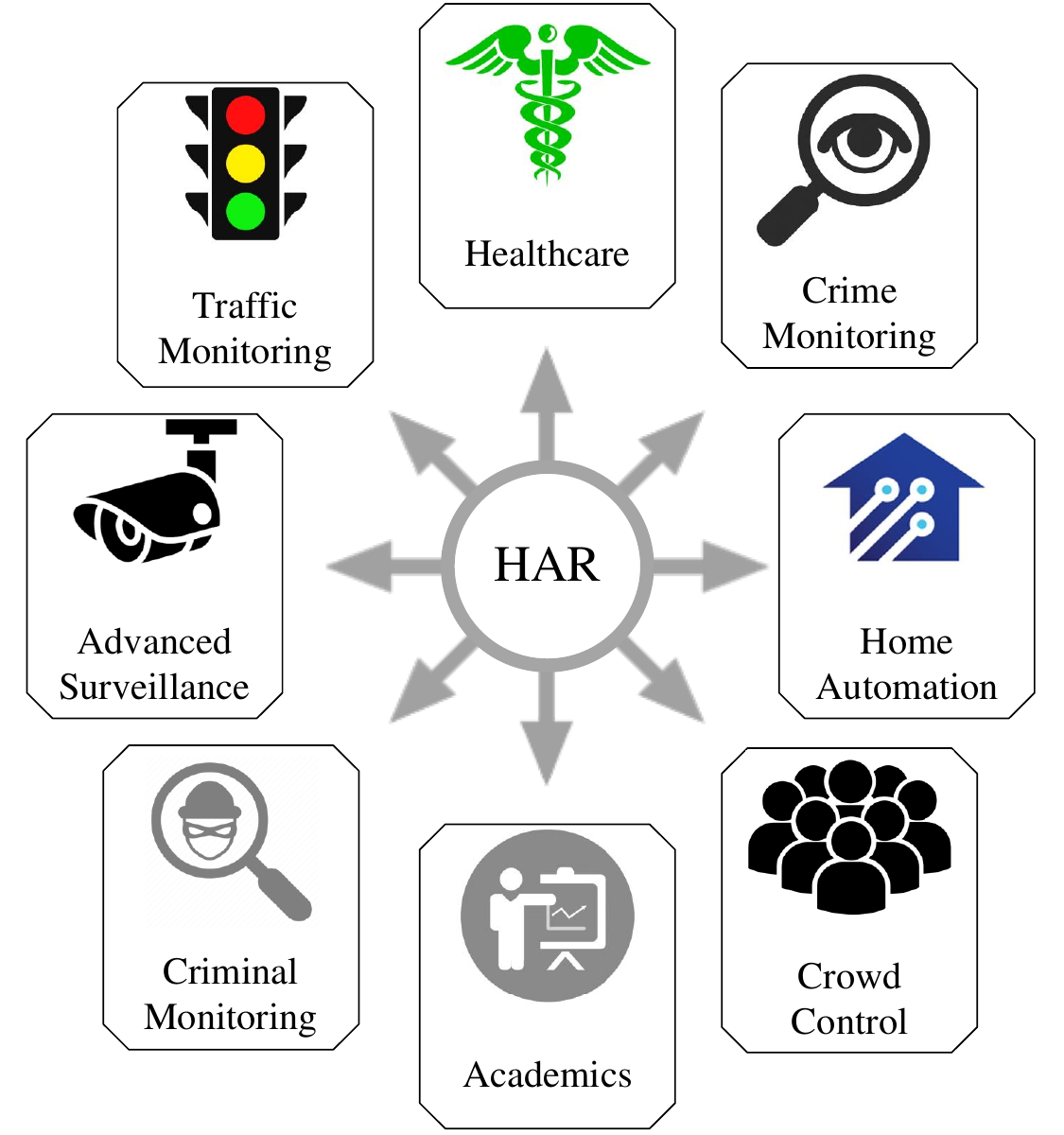}
 \caption{Future scopes of HAR }
\label{future}
\end{figure}

HAR is a provoking yet unsolved problem although great progress has been made, there remain some unsolved challenges. HAR involves tracking, detecting, and recognizing human action which leads to some challenges as subject sensitivity, activity sensitivity, inaccurate gesture detection, insufficient training data \cite{sunny2015applications}. Fig. \ref{future} indicates some of the future trends of HAR. Some of the challenges faced by finding anomalies in HAR are discussed in this section.\par

\subsection{Differentiating among the level of anomalies }
One of the greatest challenges of HAR techniques is identifying the usual and unusual activity patterns. Another major challenge is to distinguish the level between usual and unusual activity. The level of anomaly between shoplifting and robbery is different just like it is between punching and stabbing. The diversity among different types of activities like in,
\begin{itemize}
    \item Actions: Walking, running, throwing, etc.
    \item Interactions: Ranging from hugging, kissing, and shaking hands to punching and killing, interactions between objects and persons.
    
\end{itemize}
are very hard to detect. Therefore, training the models to differentiate between the anomalies has become a huge challenge for upcoming researchers working in this sector.  

\subsection{Availability of Proper Dataset}
Another major issue is the availability of labeled sets of data for model training and testing used for the detection and affirmation of anomaly sequences. Deep learning models are data-dependent models hence their accomplishment is highly influenced by the size, standards,
and features regarding the training data. Datasets of HAR include many videos divided into classes regarding the activities. Many of the datasets used in the reviewed literature like UCF101 \cite{soomro2012ucf101} and HMDB51 \cite{kuehne2011hmdb} contains basic activities such as standing up, sitting down, waving, playing musical instruments playing. But for surveillance system anomalous activity related datasets have much more impact like in CAVIAR \cite{crowley2004context}, Visor, IXMAS \cite{weinland2006free,wu2012silhouette} etc. Hence creating more datasets containing activities for surveillance systems is an important task.

\subsection{Developing Models for SBC}
Human Activity Recognition models are very power-demanding models. To process the models very powerful hardware setups with more than capable graphics processing units are needed. In the HAR system, there are multiple cameras placed in very crowded locations to gather data, and then it is processed in super-powerful computers where the most up-to-date Deep learning models run smoothly. But it is a hard task to install heavy setups in such crowded places as it is very costly and space-consuming. As in the past decade technologies in edge devices like Single Board Computers are flourishing rapidly and many developers around the world dedicating time and efforts to develop applications for these devices. But these up-to-date Deep learning models do not run properly in SBCs as they need heavy GPU to operate. Hence, there arises the need of optimizing these models. Many types of researches are being conducted to optimizing Deep learning models to run them in lower-end devices. It is a challenge to look forward to in the upcoming decade.

\subsection{Developing system for Real world}
Besides the difficulties and problems mentioned above, there are also some challenges when the system is executed in the real world such as anthropometric variation \cite{kumari2011human,jiang2012recognizing,bilinski2015video,rahmani2017learning,baradel2017pose}, multi-view variation \cite{rahmani2017learning,zhang2013cross,wang2014cross,liu2017view}, low clarity, variable lighting, weather conditions, camera motion \cite{hadfield2017hollywood,chebli2018pedestrian}, cluttered and dynamic background \cite{xu2016exploring,duckworth2016unsupervised}, input video quality and resolution, shadow and scale variation, occlusions etc \cite{jegham2020vision} or in robotic systems \cite{niloy2021critical}. Similar activities vary from person to person based on posture, motion, or appearance. Moreover, the models used in much previous literature tried to solve the HAR problem from a fixed viewpoint and their claimed success is based on that viewpoint. But when multiple viewpoints are considered the accuracy rates drop significantly because considering another viewing angle means the motion of the body, lighting, shadow, scale everything seems to be different. Hence the HAR models can not recognize the activities properly as their training dataset had not prepared them for other angles. The change in weather also creates changes in clarity and lighting which affects HAR. Researchers have been working to resolve these issues for a long time now. Using multiple synchronized dynamic cameras seems to be a promising option as it can capture from multiple viewpoints. But optimizing models according to the solution is not that easy as it generates a tremendous volume of data that increases computation time and complexity. Hence developing the method for achieving a generalized solution for HAR in the real world is a sector to be worked on in this decade.

  \section{Conclusion}
The necessity of a surveillance system using human activity recognition is skyrocketing with each day passing by as we want the world to be a safe and secured place. Despite the widespread need, HAR is yet to achieve the level to deliver the generalized performance needed for it to be used in the surveillance system in the real world scenario.
In this review, the basic solution towards HAR has been discussed and an overview of state-of-the-art supervised and unsupervised learning methods has been investigated. The methods are analyzed and compared in terms of their adopted techniques, performance, advantages, and limitations from the latest researches. Finally, a number of major challenges and possible solutions are introduced for the upcoming researches. From the review, it conceivably can be said that vision-based HAR is an uprising and prominent sector in the field of computer vision and lots of challenges remain before these methods can be used for any real-world applications.

\bibliographystyle{unsrt}  
\bibliography{references}

\end{document}